# Real-Time Traffic Density Estimation: A Hybrid Computer Vision and Predictive Machine Learning Pipeline

**Appiah Boadu Prince Kwabena**
Department of Computer Engineering
Kwame Nkrumah University of Science and Technology
Kumasi, Ghana
pkappiahboadu@st.knust.gh.edu

## Abstract

Accurate traffic density estimation from camera feeds is a prerequisite for routing, signal control, and traffic forecasting. Existing detectors count every vehicle in the frame without distinguishing active road participants from parked cars or off-road objects, producing systematically biased estimates that propagate into downstream models. Forecasting methods, including physics-informed and temporal neural approaches, assume access to clean upstream density signals, yet no real-time pipeline produces them directly from raw video under practical hardware constraints.

We introduce a two-stage framework that bridges this gap. The first stage pairs object detection and tracking with a spatial logic layer that filters detections via intersection-based reasoning against a precomputed road mask, ensuring only on-road vehicles contribute to density measurements. The second stage, a temporal estimator (LSTM, GRU, or TCN), consumes the cleaned telemetry to forecast short-horizon density. The decoupled design supports high-fidelity spatial reasoning and efficient temporal modeling across heterogeneous hardware.

We evaluate the pipeline across multiple YOLO variants and RT-DETR, benchmarking on hardware ranging from embedded systems to GPU-accelerated platforms. Results are presented as a trade-off frontier of throughput, spatial precision, and predictive accuracy using FPS, MAE in vehicle counts, and RMSE in density forecasts. The spatial logic layer significantly reduces noise-induced estimation errors, and the downstream estimator achieves robust performance under real-time constraints.

## 1 Introduction

Traffic density, defined as the number of vehicles occupying a given stretch of road at a given time, is the primary quantity that routing algorithms, signal controllers, and congestion forecasting systems consume. Because flow and average speed can both be derived from density, estimating it reliably from camera feeds is the practical foundation of any vision-based traffic intelligence system.

Two fundamental gaps in the current literature prevent this estimation from being done well. On the perception side, object detectors treat every vehicle in the frame equally, meaning that parked cars, vehicles on sidewalks or service roads, and other off-road objects all inflate the raw count. No widely adopted pipeline applies spatial reasoning to separate active on-road participants from this environmental noise, and the result is systematically biased density estimates that propagate errors into any model that consumes them. On the prediction side, forecasting methods, whether physics-informed or purely data-driven, assume they receive accurate density signals as input, yet the literature offers no upstream component that produces those signals from raw video at frame rates suitable for live deployment. The interface between perception and prediction, where noisy visual observations must be cleaned before they can be forecast, remains largely unaddressed.

Hardware introduces an additional practical constraint. Larger and more accurate detection models demand discrete GPUs, but many cities, particularly in developing regions, can only afford embedded edge devices for wide-area deployment. A pipeline that runs exclusively on high-end hardware is therefore not deployable where traffic monitoring is needed most. Any system intended for real-world adoption must expose useful operating points across a range of compute budgets, from lightweight embedded boards to GPU-accelerated servers.

This paper introduces a two-stage, hardware-aware framework that addresses both gaps simultaneously. In the first stage, a detector from the YOLO family or RT-DETR is paired with an object tracker and a spatial logic layer. The spatial logic layer compares each detected bounding box against a static road mask, derived from training data annotated on the Roboflow platform [1], and discards any detection whose overlap with the drivable area falls below a configurable threshold. The output is a per-frame telemetry stream containing on-road vehicle counts, occupancy ratios, and movement statistics, all free of the off-road noise that contaminates conventional detector output. In the second stage, a temporal estimator implemented as an LSTM, GRU, or Temporal Convolutional Network consumes this cleaned stream and forecasts density over a short horizon. The decoupled design allows the perception and prediction components to be developed, optimized, and deployed independently.

We evaluate the pipeline across six detector families, YOLOv5 [2], YOLOv8 [3], YOLOv9 [4], YOLOv10 [5], YOLO11 [6], and RT-DETR [7], selected to demonstrate that the architecture is model-agnostic; when these models are not individually optimized, architectures of comparable scale converge to similar accuracy, reinforcing the point that the spatial logic layer, not the choice of detector, is the distinguishing component. Results are reported as a trade-off frontier of throughput versus error, using frames per second, mean absolute error on vehicle counts, and RMSE on density forecasts, so that a deployment engineer can match a given hardware budget to the accuracy it can sustain.

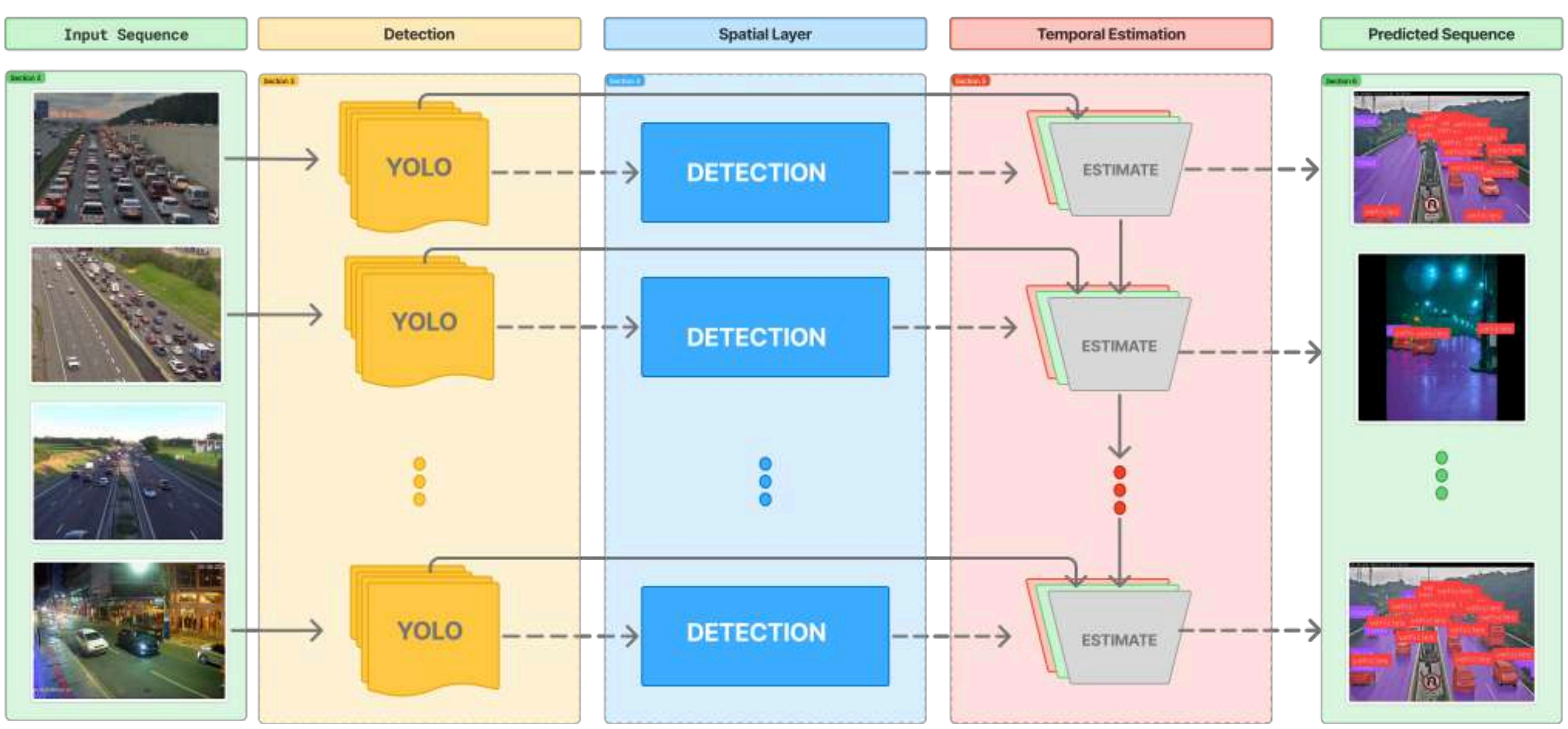

Figure 1: The hybrid architecture bridging raw video input via the Spatial Logic Layer to the Downstream ML Estimator.

The remainder of the paper is organized as follows. Section 2 reviews related work on visual perception, hardware-aware estimation, and physics-informed prediction. Section 3 states the problem formally. Section 4 presents the method, Section 5 the experimental setup, and Section 6 the results. Section 7 concludes.

## 2 Related Work

Prior work in traffic density estimation can be organized into three threads: visual perception of traffic, hardware-constrained estimation on embedded devices, and physics-informed prediction. The first two address the input side of the problem, that is, how to read the state of the road from a camera. The third addresses the output side, how to forecast where the traffic state is heading. Existing systems tend to operate entirely on one side or the other, and the gap our work targets is the interface between them.

### 2.1 Visual Perception

Sadik et al. [8] evaluate YOLOv8 and RT-DETR on urban traffic scenes and show that modern single-stage detectors remain accurate on small objects and under poor lighting, conditions that older detector generations handled badly. Their results establish that single-stage detectors are now strong enough to serve as the primary sensor for real-time traffic monitoring. Nubert et al. [9] take a coarser approach, classifying entire camera frames into one of five density bands using a transfer-learned InceptionV3 and arguing that the translation invariance of CNNs makes this robust to the minor camera shifts typical of public deployments. Both studies confirm that pixel data alone supports useful density labels, but neither separates on-road traffic from off-road vehicles, and neither attempts to filter parked cars from the count. That filtering gap, where raw detections must be spatially validated against a known road boundary before they can be trusted, is precisely what our spatial logic layer is designed to address.

### 2.2 Hardware-Aware Estimation

Zoysa and Munasinghe [10] demonstrate that perception can be moved onto extremely cheap hardware. Rather than running a full detector, they extract edge pixels from a fixed road region and map the resulting intensity to a vehicle count via a small DNN, all on a Raspberry Pi. Their work shows that GPU-class hardware is not a prerequisite for traffic estimation when the upstream representation is chosen carefully. We adopt the same constraint throughout our design: any pipeline intended for wide deployment must expose a useful operating point on embedded targets, not only on workstation-class accelerators.

### 2.3 Physics-Informed Prediction

Where perception models read the present state of the road, physics-informed neural networks (PINNs) attempt to forecast the next interval by embedding traffic-flow equations directly into the learning objective. Pereira et al. [11] incorporate the macroscopic Traffic Reaction Model into a recurrent network, producing a grey-box predictor that retains physical interpretability and tolerates noisy or partially missing sensor data. Wilkman et al. [12] extend this idea to an online setting in which the model trains iteratively on incoming probe-vehicle data and adapts to changing regime parameters such as free-flow speed. Both groups report a recurring difficulty: the gradients from the data-fitting loss and the physics-residual loss often conflict, which slows convergence and complicates training.

### 2.4 Optimization for Hybrid Models

Wang et al. [13] tackle this loss-conflict problem directly with a multi-task optimization framework that trains a PINN jointly on speed and density. Auxiliary tasks supply trans-ferable gradient signal, and the reported effect is faster convergence and better accuracy on the sparse data typical of real-world traffic sensors.

### 2.5 Positioning

Across these threads, the literature divides into a perception side that produces increasingly accurate but noise-contaminated readings on ever-cheaper hardware, and a prediction side that models flow dynamics correctly but assumes the perception side has already produced clean inputs. The two halves have not been joined under live, real-time constraints. Our contribution sits at that interface. The spatial logic layer filters perception output by comparing bounding boxes against a precomputed road mask before it reaches the predictor, and the predictor itself is a temporal model (LSTM, GRU, or TCN) rather than a PINN. We trade strict physical interpretability for more stable training and a real-time inference profile that is compatible with embedded deployment, the same practical constraint that motivates [10].

## 3 Background: Problem Setting

Traffic density at time $t$ on a road segment of length $L$ is defined as

$$\rho(t) = \frac{N(t)}{L}, \tag{1}$$

where $N(t)$ is the number of vehicles on the segment. In a camera-based system, $L$ is not measured directly; instead, the camera observes a fixed field of view in which the road occupies a known pixel region. We therefore reformulate density in terms of pixel occupancy. Let $M_R$ denote the binary road mask (1 for drivable pixels, 0 otherwise) and let $M_V^i$ denote the pixel footprint of the $i$-th detected vehicle. The on-road occupancy ratio is

$$O(t) = \frac{|\bigcup_{i \in \mathcal{S}} M_V^i \cap M_R|}{|M_R|}, \tag{2}$$

where $\mathcal{S}$ is the set of vehicles classified as on-road by the spatial logic layer. A vehicle $i$ is included in $\mathcal{S}$ if and only if its box-road overlap ratio exceeds a threshold $\tau$:

$$r_i = \frac{|B_i \cap M_R|}{|B_i|} \geq \tau, \tag{3}$$

where $B_i$ is the set of pixels inside the bounding box of vehicle $i$. This formulation intentionally uses the fraction of the box that falls on road, not a symmetric IoU, because the question we need to answer is "how much of this vehicle is on the road," not "how similar are the two regions."

Given the filtered set $\mathcal{S}$, the pipeline extracts a ten-dimensional feature vector $\boldsymbol{f}(t)$ per frame comprising vehicle count, occupancy ratio, mean speed, mean direction, density, flow, congestion index, stopped-vehicle ratio, speed variance, and direction variance. A temporal estimator then maps a sliding window of these features to a short-horizon density forecast:

$$\hat{\boldsymbol{f}}(t+1 : t+H) = g(\boldsymbol{f}(t-W+1 : t); \theta), \tag{4}$$

where $W$ is the lookback window length, $H$ is the forecast horizon, and $g(\cdot;\theta)$ is a learned model (LSTM, GRU, or TCN) with parameters $\theta$.

## 4 Method

The system is organized as a two-stage pipeline. Stage 1 (perception) transforms raw video into a clean per-frame telemetry stream. Stage 2 (prediction) consumes that stream and forecasts density over a short horizon. We describe each stage in detail below.

### 4.1 Stage 1: Data Preparation

Training data was sourced from the Roboflow public asset library [1] using search filters for road, traffic jam, and vehicle imagery. The resulting dataset comprises 5,716 annotated frames with two object classes: *road* and *vehicles*. Annotations are in COCO bounding-box format. We split the data at the frame level into training (4,001 frames, 70%), validation (857 frames, 15%), and test (858 frames, 15%) partitions, then convert to YOLO-format label files for compatibility with the Ultralytics training pipeline.

From the road-class annotations we also derive a static binary road mask $M_R$ by accumulating per-frame road regions and applying morphological closing and opening to remove noise. This mask is used at inference time by the spatial logic layer and does not change between frames.

### 4.2 Stage 2: Detection Model Training

We train ten detector variants spanning six architectural families, as summarized in Table 1. All models are initialized from pretrained COCO weights and fine-tuned on our dataset using the AdamW optimizer with an initial learning rate of $10^{-3}$, cosine-annealed to $10^{-2} \times \mathrm{lr}_0$. Training uses a batch size of 16, an input resolution of $640 \times 640$, mosaic augmentation, and early stopping with a patience of 20 epochs. Table 2 lists the full set of shared hyperparameters.

Table 1: Detection models evaluated. Parameter counts and weight sizes are for the fine-tuned checkpoints.

| Model | Family | Params | Weights (MB) | Epochs |
|---|---|---|---|---|
| YOLOv5n | YOLOv5 | 2.5 M | 5.0 | 100 |
| YOLOv5s | YOLOv5 | 9.1 M | 17.7 | 100 |
| YOLOv5m | YOLOv5 | 25.1 M | 48.2 | 100 |
| YOLOv5l | YOLOv5 | 53.2 M | 101.8 | 100 |
| YOLOv8n | YOLOv8 | 3.0 M | 6.0 | 100 |
| YOLOv8s | YOLOv8 | 11.1 M | 21.5 | 100 |
| YOLOv9c | YOLOv9 | 25.5 M | 49.2 | 50 |
| YOLOv10m | YOLOv10 | 16.5 M | 31.9 | 50 |
| YOLO11m | YOLO11 | 20.1 M | 38.6 | 50 |
| RT-DETR-l | RT-DETR | 32.8 M | 63.2 | 100 |

Table 2: Shared training hyperparameters across all detector variants.

| Hyperparameter | Value |
|---|---|
| Optimizer | AdamW |
| Initial learning rate | $10^{-3}$ |
| LR schedule | Cosine annealing to $0.01 \times \mathrm{lr}_0$ |
| Batch size | 16 |
| Image size | $640 \times 640$ |
| Warmup epochs | 5 |
| Early stopping patience | 20 |
| Weight decay | $5 \times 10^{-4}$ |
| Mosaic augmentation | Enabled |
| Horizontal flip | 0.5 |
| Erasing augmentation | 0.4 |
| Mixed precision (AMP) | Enabled |
| Pretrained weights | COCO |

Figure 2 shows the training and validation loss curves across all models. All variants converge within their allocated epoch budgets, with the larger models (YOLOv5l, YOLOv9c) achieving marginally lower final validation loss. Figure 3 shows the evolution of mAP@50 and mAP@50-95 during training; the metrics plateau in the final third of training for all models, confirming that early stopping with patience 20 is sufficient.

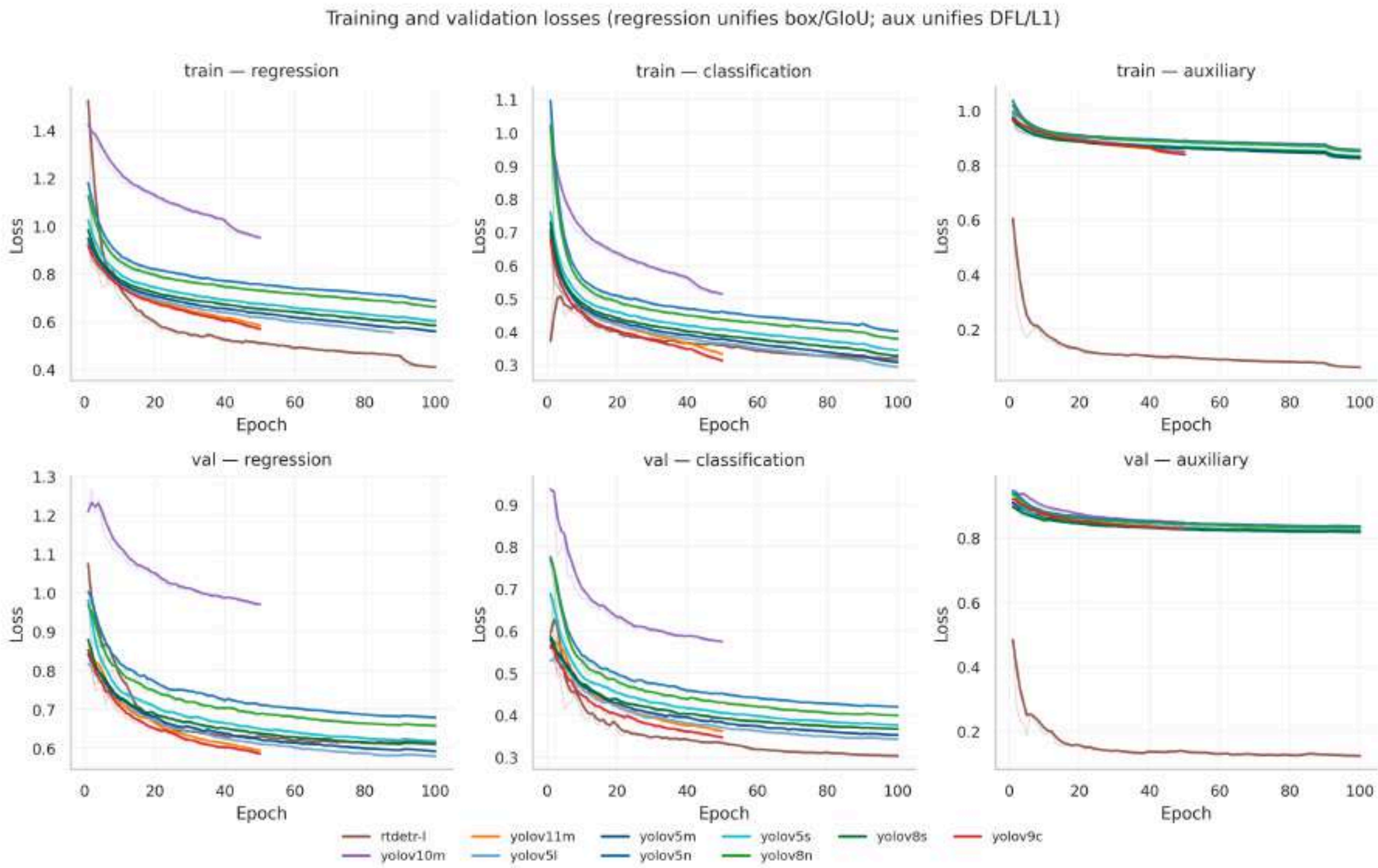

Figure 2: Training and validation loss curves for all ten detector variants.

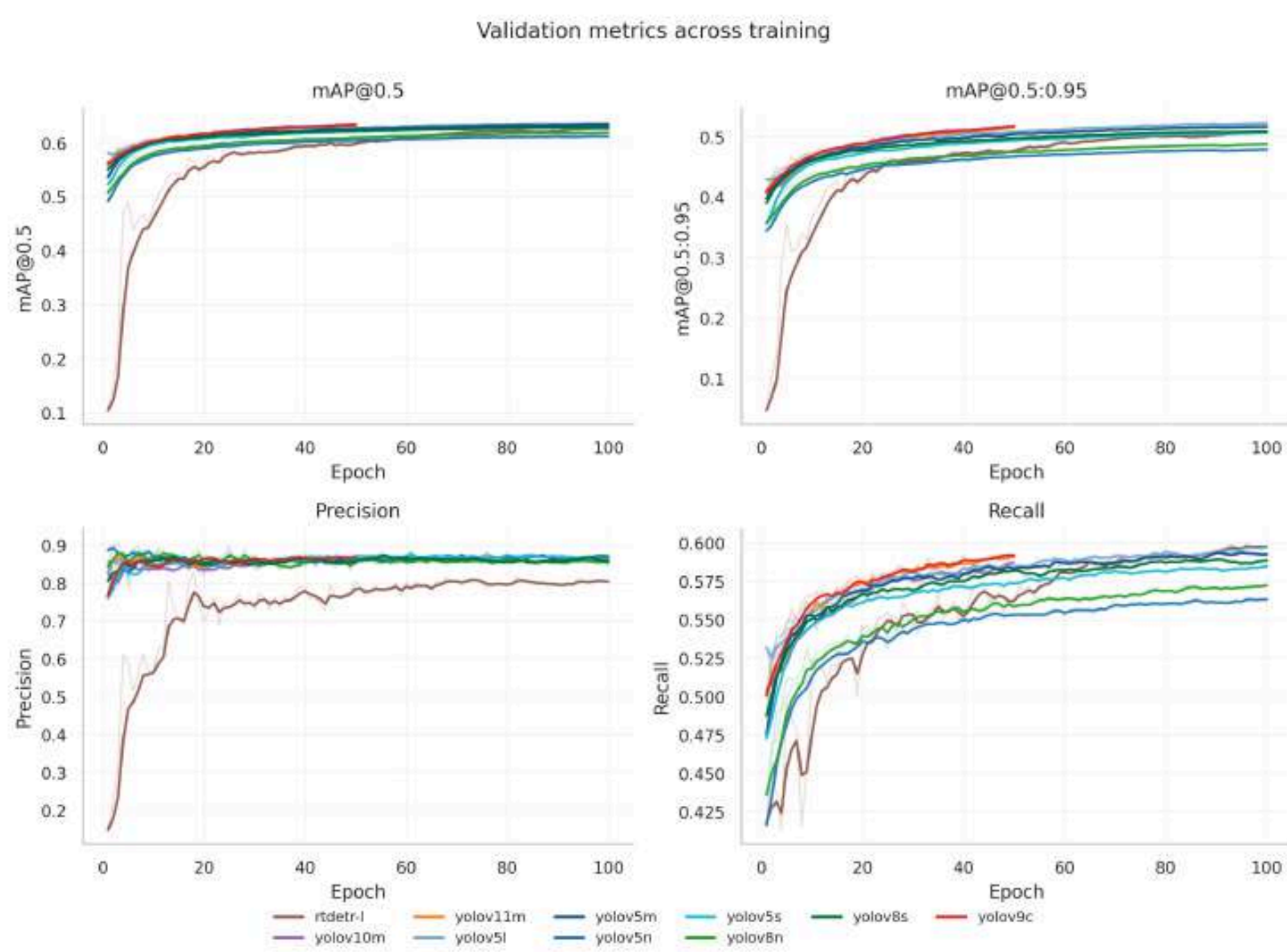

Figure 3: Evolution of mAP@50 and mAP@50-95 during training across all models.

### 4.3 Stage 3: Object Tracking

During inference, the detector is paired with an object tracker, either BoT-SORT [14] or ByteTrack [15], via the Ultralytics built-in tracking API. The tracker assigns persistent integer IDs to each detected vehicle across frames, enabling the computation of per-vehicle movement vectors. For each tracked vehicle, we compute the velocity vector $\boldsymbol{v} = \boldsymbol{c}_t - \boldsymbol{c}_{t-1}$ from consecutive centroid positions, the speed $s = \|\boldsymbol{v}\|$, and the direction $\varphi = \operatorname{atan2}(v_y, v_x)$ in degrees.

To prevent redundant counting when vehicles re-enter the frame or when IDs are reassigned, we implement a counting module based on two complementary mechanisms: line-crossing detection using the cross product of the track's displacement against a counting line, and zone-entry detection using a ray-casting point-in-polygon test. Each track is counted at most once per line or zone, regardless of how many frames it remains visible.

### 4.4 Stage 4: Spatial Logic Layer

The spatial logic layer is the core filtering component that distinguishes our pipeline from conventional detection-based density estimation. Given the set of tracked vehicles from Stage 3 and the precomputed road mask $M_R$, the filter classifies each vehicle as on-road or off-road using the box-road overlap ratio defined in the Background section.

For each vehicle $i$ with bounding box $B_i = [x_1, y_1, x_2, y_2]$, we compute

$$r_i = \frac{\sum_{p \in B_i} M_R(p)}{|B_i|}. \tag{5}$$

If $r_i \geq \tau$ (default $\tau = 0.5$), the vehicle is classified as on-road and included in the filtered set $\mathcal{S}$. An optional minimum-speed gate further excludes stationary vehicles (speed below a configurable threshold) even if they overlap the road, which is useful for filtering vehicles stopped at the roadside.

The filter then computes the on-road occupancy ratio $O(t)$ as the fraction of road pixels covered by the union of on-road bounding boxes. The output is a `FilteredResult` containing the list of on-road vehicles with their track IDs, bounding boxes, centroids, speeds, directions, and the aggregate occupancy ratio and vehicle count.

### 4.5 Stage 5: Feature Extraction

From each `FilteredResult`, we extract a ten-dimensional feature vector $\boldsymbol{f}(t) \in \mathbb{R}^{10}$:

Table 3: The ten per-frame features extracted from the spatial logic layer output.

| Index | Feature | Description |
|---|---|---|
| 0 | vehicle_count | Number of on-road vehicles |
| 1 | occupancy_ratio | Fraction of road pixels covered |
| 2 | mean_speed | Mean speed of on-road vehicles (px/frame) |
| 3 | mean_direction | Mean heading in degrees |
| 4 | density | vehicle_count / road_area |
| 5 | flow | Vehicles crossing a counting line per frame |
| 6 | congestion_index | $O(t) \times (1 - \bar{s}/s_{\max})$, clipped to $[0, 1]$ |
| 7 | stopped_ratio | Fraction of vehicles with speed $< 1$ px/frame |
| 8 | speed_variance | Variance of on-road vehicle speeds |
| 9 | direction_variance | Variance of on-road vehicle headings |

The congestion index deserves particular note: it combines occupancy with the inverse of normalized speed, so a road that is both highly occupied and slow-moving scores close to 1.0, while a road that is occupied but free-flowing scores lower. This provides a single scalar summary of congestion severity.

### 4.6 Stage 6: Temporal Density Estimation

The feature stream from Stage 5 is consumed by a temporal estimator that forecasts density over a short horizon. We implement three architectures:

- **LSTM** [16]: A multi-layer Long Short-Term Memory network with a fully connected head. Configurable hidden size (default 64), number of layers (default 2), dropout, and optional bidirectional mode.
- **GRU** [17]: A multi-layer Gated Recurrent Unit with the same head structure. GRUs have fewer parameters than LSTMs for the same hidden size, offering a speed advantage at comparable accuracy.
- **TCN** [18]: A Temporal Convolutional Network with dilated causal convolutions and residual connections. Each temporal block applies two weight-normalized 1D convolutions with exponentially increasing dilation, followed by ReLU activation and dropout. The receptive field grows exponentially with depth, allowing the network to capture long-range dependencies without recurrence.

All three architectures take as input a window of $W$ feature vectors (default $W = 10$) and produce a forecast of $H$ steps ahead (default $H = 5$). Features are normalized before input using one of three methods: standard (z-score), min-max, or robust (median/IQR). The normalizer is fit on the training set and applied identically at inference time.

### 4.7 Inference Pipeline Architecture

At inference time, the six stages above execute as a multi-threaded pipeline. Each stage runs in its own daemon thread, connected to the next by bounded queues with a configurable overflow policy (default: drop oldest frame). This design ensures that a slow downstream stage does not block the video reader, and that the system degrades gracefully under load by dropping frames rather than accumulating unbounded latency.

A fixed-FPS controller paces the pipeline. If processing is ahead of schedule, the controller sleeps to maintain the target frame rate. If processing falls behind, it skips frames to prevent queue buildup. The controller tracks realized FPS over a sliding window and reports processing statistics at configurable intervals.

The output is an annotated video with a heads-up display showing the road mask overlay, tracked bounding boxes with IDs and speeds, and a panel reporting real-time vehicle count, occupancy ratio, congestion index, and density forecast. A parallel CSV writer logs the full ten-feature telemetry stream for offline analysis.

## 5 Experimental Setup

### 5.1 Hardware and Software

All training and benchmarking experiments were conducted on a single NVIDIA GPU using CUDA, with the Ultralytics framework (v8.2+) for detection, tracking, and model export. The inference pipeline is implemented in Python using PyTorch, OpenCV, and NumPy. Latency measurements use `torch.cuda.synchronize()` barriers to ensure accurate GPU timing.

### 5.2 Benchmark Protocol

Each trained model is evaluated in two phases. First, we run the Ultralytics validation routine on the held-out test split (858 images) to obtain detection metrics: precision, recall, mAP@50, mAP@50-95, F1, and per-class average precision. Second, we run a dedicated latency loop on 300 frames decoded from a representative video clip. The loop includes 10 warm-up iterations (excluded from timing) followed by 100 timed iterations with CUDA synchronization, yielding a latency distribution from which we report mean, median, P95, P99, and derived FPS.

This two-phase protocol evaluates the detector in isolation, without tracker or spatial logic overhead, so that the reported numbers reflect the model itself and can be compared fairly across families.

### 5.3 Estimator Training Protocol

To evaluate the temporal estimation stage, we collect telemetry from a live Caltrans HLS traffic camera stream using the best-performing detector from the benchmark above (YOLOv5l) paired with BoT-SORT tracking. Coupling the estimator to the highest-accuracy detector minimizes the upstream noise floor and lets us isolate the temporal model's contribution. The pipeline runs for four hours, producing a per-frame CSV of the ten features described in Table 3. We train three estimator architectures — LSTM, GRU, and TCN — on this telemetry using the hyperparameters summarized in Table 4. Each CSV is split temporally (80% train, 20% validation) with no cross-stream splicing, preserving the causal ordering of the time series. Features are z-score normalized using statistics computed on the training partition only.

Evaluation uses a held-out stream from a different camera, yielding 29,243 sliding windows. Predictions are inverse-transformed to original feature units before computing metrics, so that MSE, MAE, RMSE, and $R^2$ reflect real-world magnitudes rather than normalized space. In addition to this offline evaluation, we run the full six-stage pipeline live on a third, previously-unseen Caltrans stream with each estimator in turn (Section 6.7), so that the reported metrics are corroborated by the system's actual end-to-end behaviour under the multi-threaded inference path.

Table 4: Shared hyperparameters for all three temporal estimator architectures.

| Hyperparameter | Value |
|---|---|
| Architectures | LSTM, GRU, TCN |
| Hidden size | 128 |
| Layers / blocks | 2 (LSTM, GRU); 3 blocks (TCN) |
| Dropout | 0.2 |
| Optimizer | Adam |
| Learning rate | $10^{-3}$ |
| LR schedule | ReduceLROnPlateau (factor 0.5, patience 10) |
| Gradient clipping | Max norm 1.0 |
| Batch size | 64 |
| Max epochs | 200 |
| Early stopping patience | 30 |
| Normalization | Standard (z-score) |
| Input window ($W$) | 10 frames |
| Forecast horizon ($H$) | 5 steps |
| Loss function | MSE |

# 6 Results and Discussion

## 6.1 Detection Accuracy

Table 5 summarizes the detection accuracy and throughput of all ten models on the test split. The top-performing model by mAP@50-95 is YOLOv5l (0.524), followed closely by YOLOv9c (0.521) and YOLOv10m (0.520). The spread across all ten models is narrow: the gap between the best and worst mAP@50-95 is only 0.043, confirming that unoptimized models of comparable training regimes converge to similar accuracy and that the choice of detector family is less important than the downstream spatial filtering.

Table 5: Benchmark results on the test split (858 images). Models ranked by mAP@50:95. Latency is the mean over 100 timed iterations with CUDA synchronization.

| Model | @50 | @50:95 | Precision | Recall | F1 | FPS | Lat. (ms) |
|---|---|---|---|---|---|---|---|
| YOLOv5l | 0.636 | 0.524 | 0.875 | 0.594 | 0.708 | 56 | 17.9 |
| YOLOv9c | 0.634 | 0.521 | 0.861 | 0.594 | 0.703 | 64 | 15.7 |
| YOLOv10m | 0.634 | 0.520 | 0.865 | 0.586 | 0.698 | 95 | 10.5 |
| YOLOv5m | 0.634 | 0.520 | 0.869 | 0.592 | 0.704 | 92 | 10.8 |
| YOLO11m | 0.633 | 0.519 | 0.863 | 0.593 | 0.703 | 87 | 11.5 |
| RT-DETR-l | 0.626 | 0.512 | 0.816 | 0.595 | 0.688 | 30 | 33.6 |
| YOLOv8s | 0.630 | 0.511 | 0.870 | 0.587 | 0.701 | 178 | 5.6 |
| YOLOv5s | 0.629 | 0.508 | 0.879 | 0.584 | 0.702 | 184 | 5.4 |
| YOLOv8n | 0.619 | 0.487 | 0.862 | 0.575 | 0.690 | 258 | 3.9 |
| YOLOv5n | 0.615 | 0.481 | 0.861 | 0.568 | 0.685 | 252 | 4.0 |

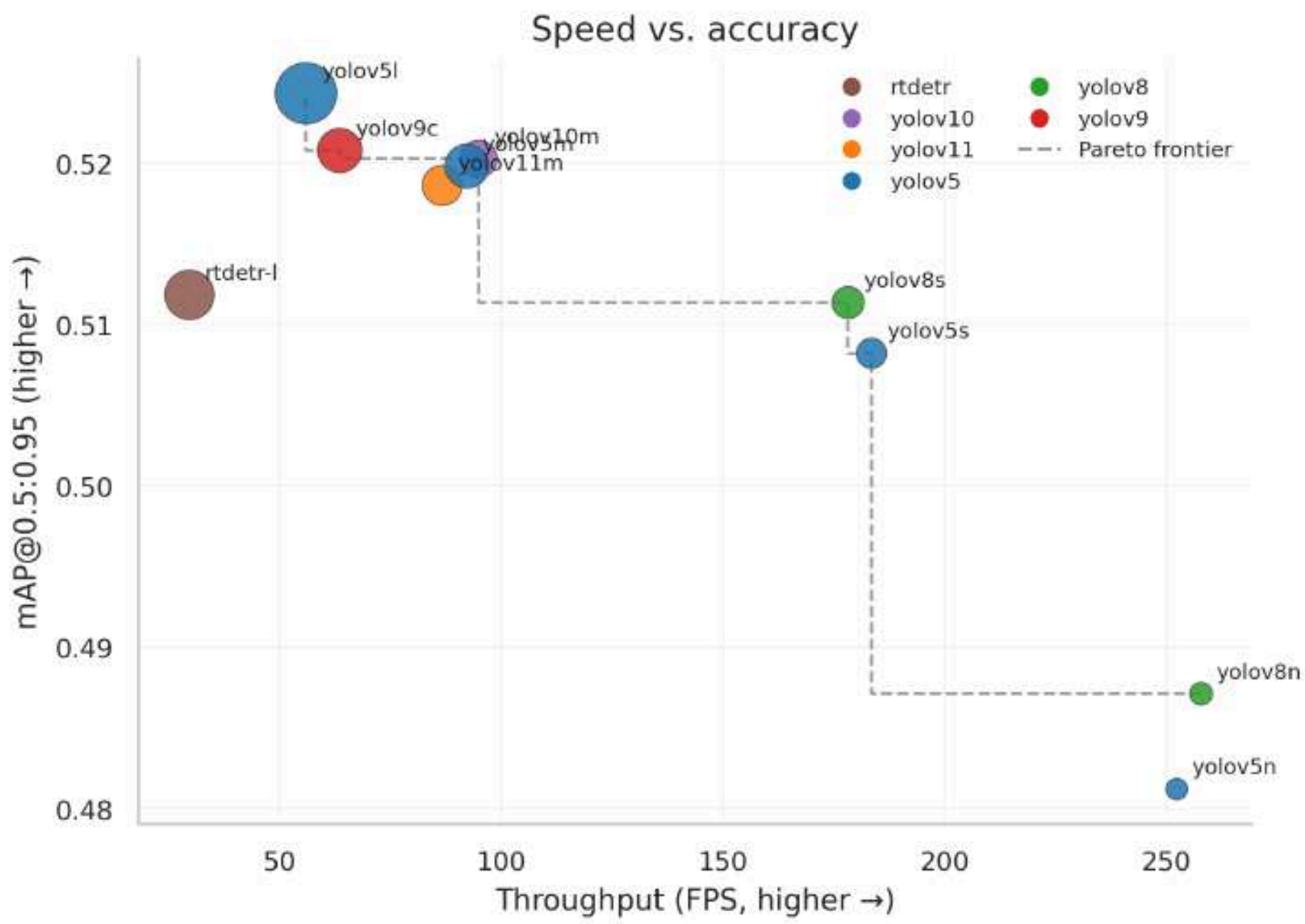

Figure 4: Speed-accuracy trade-off across all detector variants. The Pareto frontier runs from YOLOv5n/YOLOv8n (fastest, lowest accuracy) through YOLOv8s/YOLOv5s (balanced) to YOLOv5l (most accurate, slowest among YOLO variants). RT-DETR-l is dominated: it is both slower and less accurate than the YOLO mid-range.

### 6.2 Speed-Accuracy Trade-off

Figure 4 plots the speed-accuracy frontier. A clear Pareto structure emerges. At the lightweight end, YOLOv5n and YOLOv8n exceed 250 FPS with mAP@50-95 around 0.48, making them suitable for embedded deployment where throughput matters more than marginal accuracy. In the mid-range, YOLOv8s and YOLOv5s achieve 178–184 FPS with mAP@50-95 above 0.50, offering an attractive balance. At the heavy end, YOLOv5l reaches the highest accuracy (0.524) at 56 FPS, still comfortably real-time for most applications. RT-DETR-l is a notable outlier: despite being a transformer-based architecture, it achieves lower accuracy than the YOLO mid-range while running at only 30 FPS, making it Pareto-dominated in this benchmark.

Figure 5 shows the relationship between model size and accuracy. Accuracy scales sub-linearly with parameter count: YOLOv5l (53M parameters) gains only 0.043 mAP@50-95 over YOLOv5n (2.5M), a 21× increase in parameters for less than 9% relative improvement. This diminishing return reinforces the case for deploying smaller models when hardware is constrained.

Figure 6 and Figure 7 provide complementary views of throughput. The FPS bar chart highlights the order-of-magnitude gap between the nano-class models (250+ FPS) and RT-DETR-l (30 FPS). The latency box plot reveals that the YOLO variants exhibit tight, low-variance latency distributions, while RT-DETR-l shows both higher median latency and wider spread, making its frame-to-frame timing less predictable — an important consideration for real-time pipelines that rely on consistent per-frame budgets.

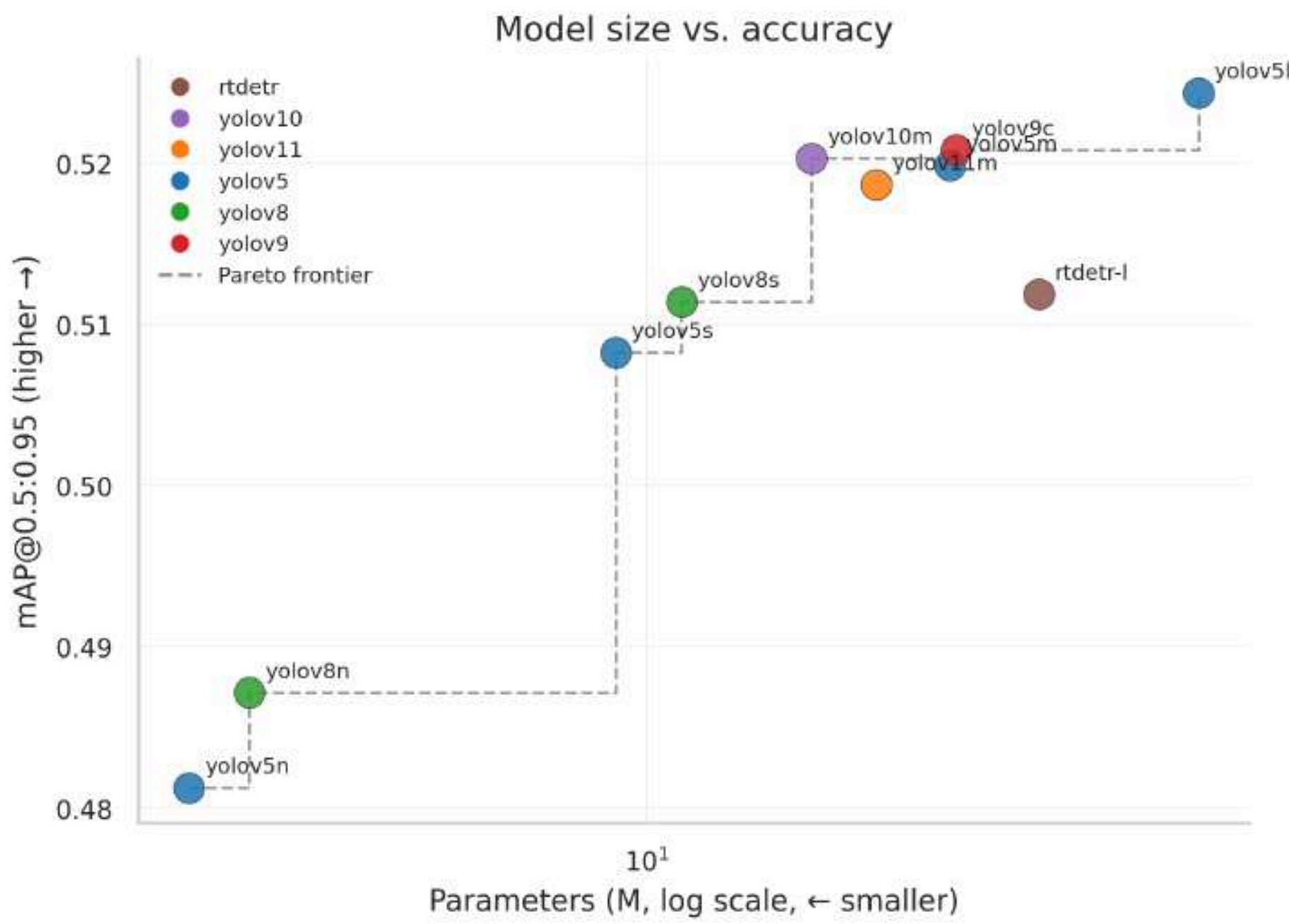

Figure 5: Model size (parameters) versus mAP@50-95. Accuracy scales sub-linearly with model size; the largest model (YOLOv5l, 53M params) gains only 0.043 mAP@50-95 over the smallest (YOLOv5n, 2.5M params).

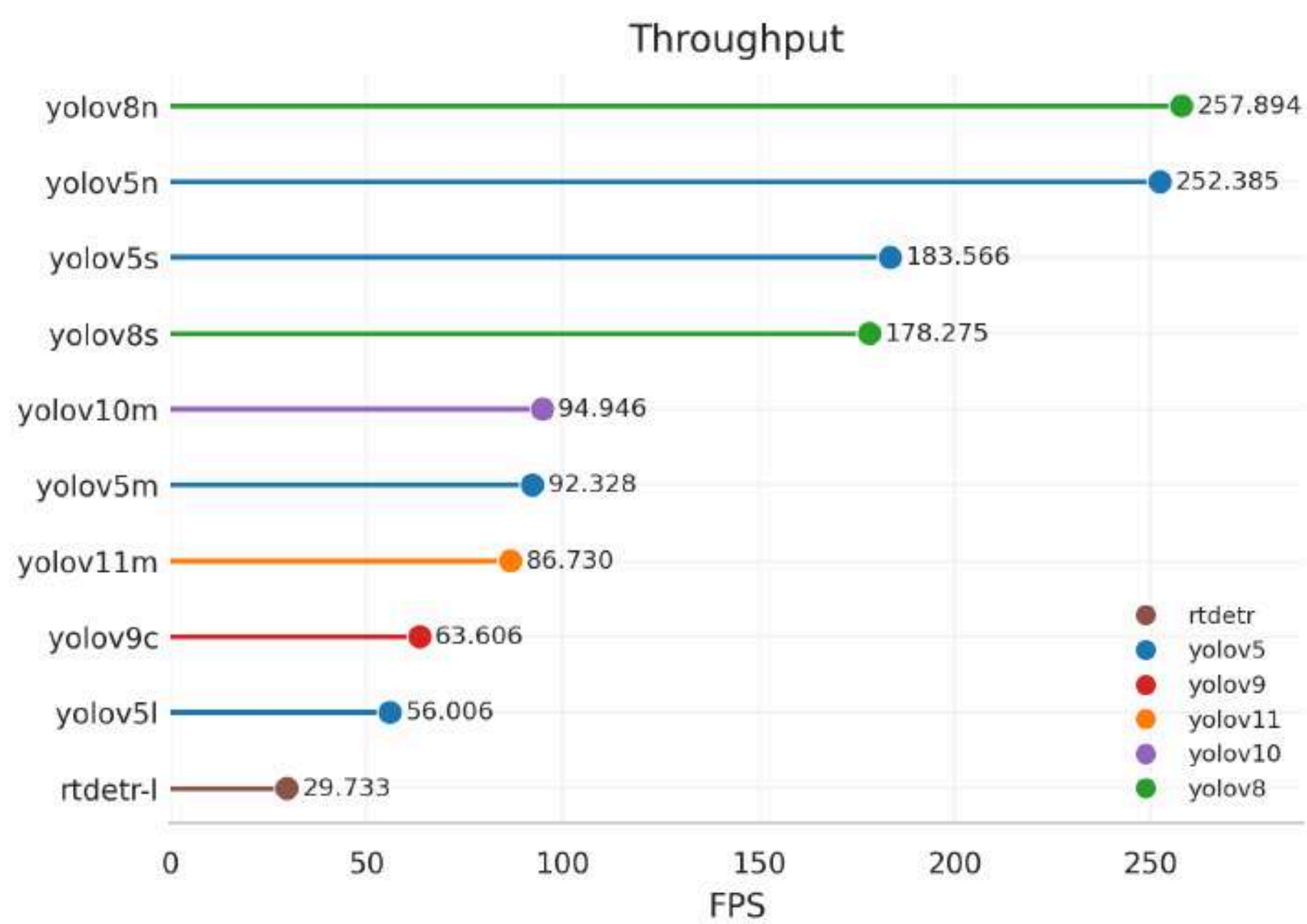

Figure 6: Throughput (FPS) comparison across all detector variants.

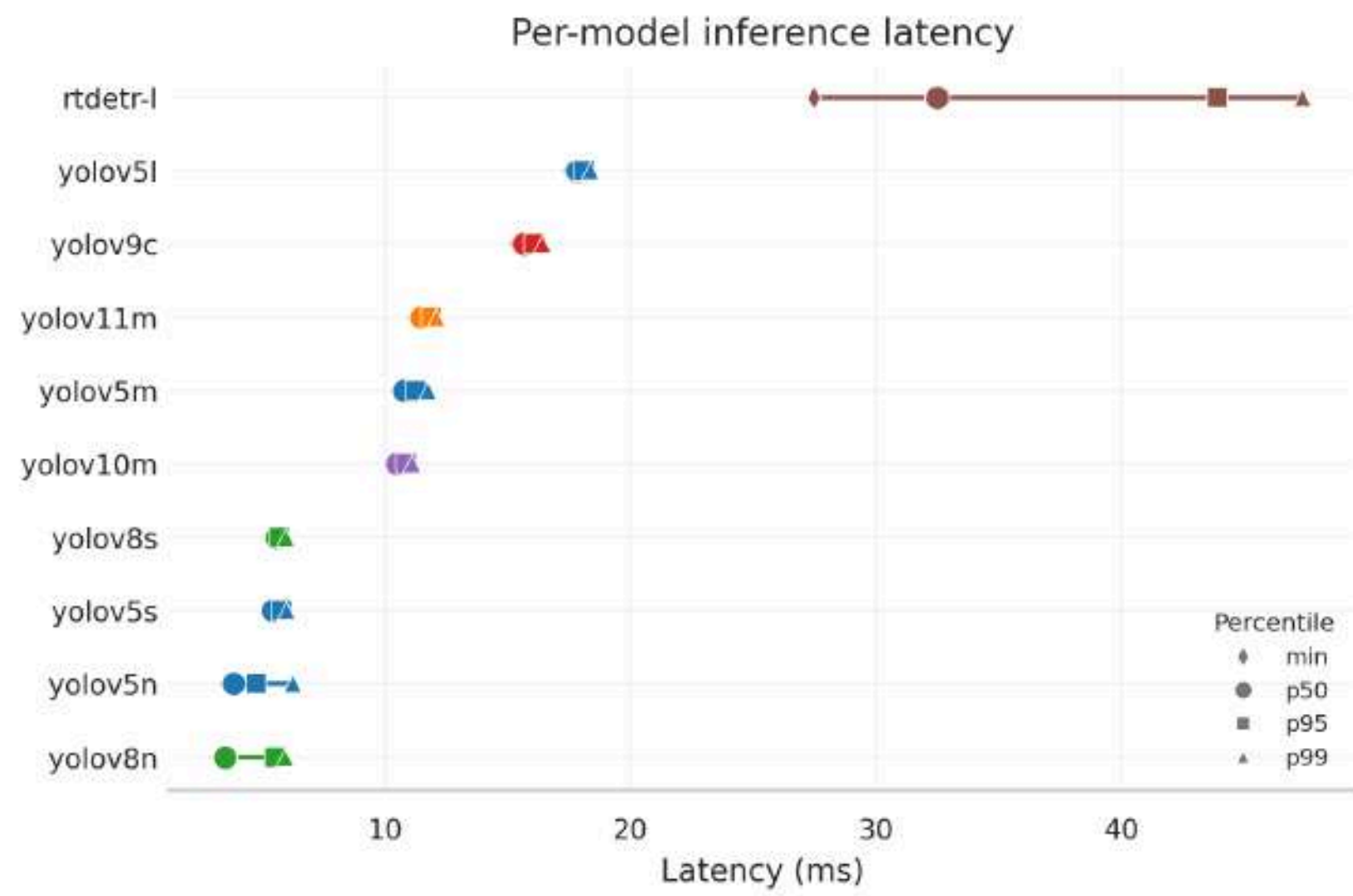

Figure 7: Latency distribution (ms per frame) across all detector variants. Box plots show median, interquartile range, and P95/P99 whiskers.

### 6.3 Per-Class Analysis

Figure 8 reveals a striking asymmetry between the two classes. The *vehicles* class achieves AP@50 between 0.94 and 0.96 across all models, indicating that vehicle detection is effectively solved for this dataset. The *road* class, by contrast, achieves AP@50 between only 0.29 and 0.31. This disparity is expected for three reasons. First, road regions are large, amorphous areas rather than compact objects, making bounding-box representation inherently imprecise. Second, road annotations tend to be partial, covering only the visible drivable surface rather than the full extent of the road. Third, the evaluation metric penalizes large bounding boxes that extend beyond the annotated region, even when the prediction is semantically correct. For our pipeline, this low road AP is not a practical concern: the road mask is precomputed once during data preparation and used as a static binary filter at inference time, so detection-time road localization accuracy does not affect the density estimate.

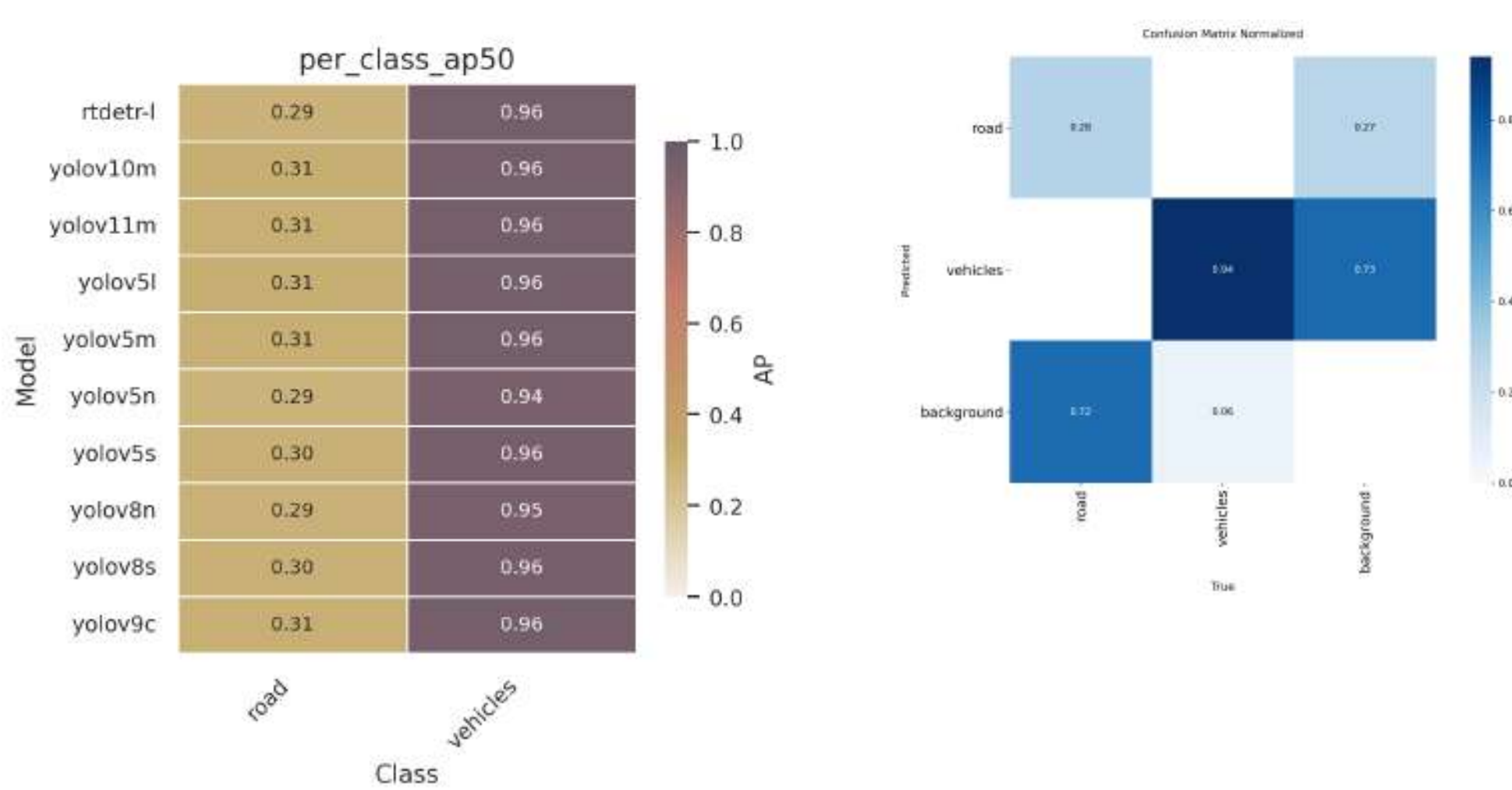

Figure 8: Left: Per-class AP@50 across all models. Vehicle detection is consistently strong (0.94–0.96); road detection is weaker (0.29–0.31). Right: Normalized confusion matrix for YOLOv5l on the test split.

Figure 8 (right) confirms this pattern for the best model (YOLOv5l): vehicle predictions are concentrated on the diagonal with minimal confusion, while road predictions show a higher rate of misclassification against the background class.

### 6.4 Training Dynamics

Figure 10 presents the full training curves for all ten models. Several patterns are consistent across families. Validation loss decreases monotonically for the first 60–80% of training, then plateaus. Precision peaks early (often within the first 5 epochs) and remains stable, while recall improves gradually throughout training. The mAP metrics follow the recall trajectory, confirming that recall is the binding constraint on overall performance. Figure 9 shows the cosine-annealed learning rate schedule shared by all models; the smooth decay from $10^{-3}$ to $10^{-5}$ avoids the abrupt drops of step-based schedules and contributes to the stable convergence observed across all families.

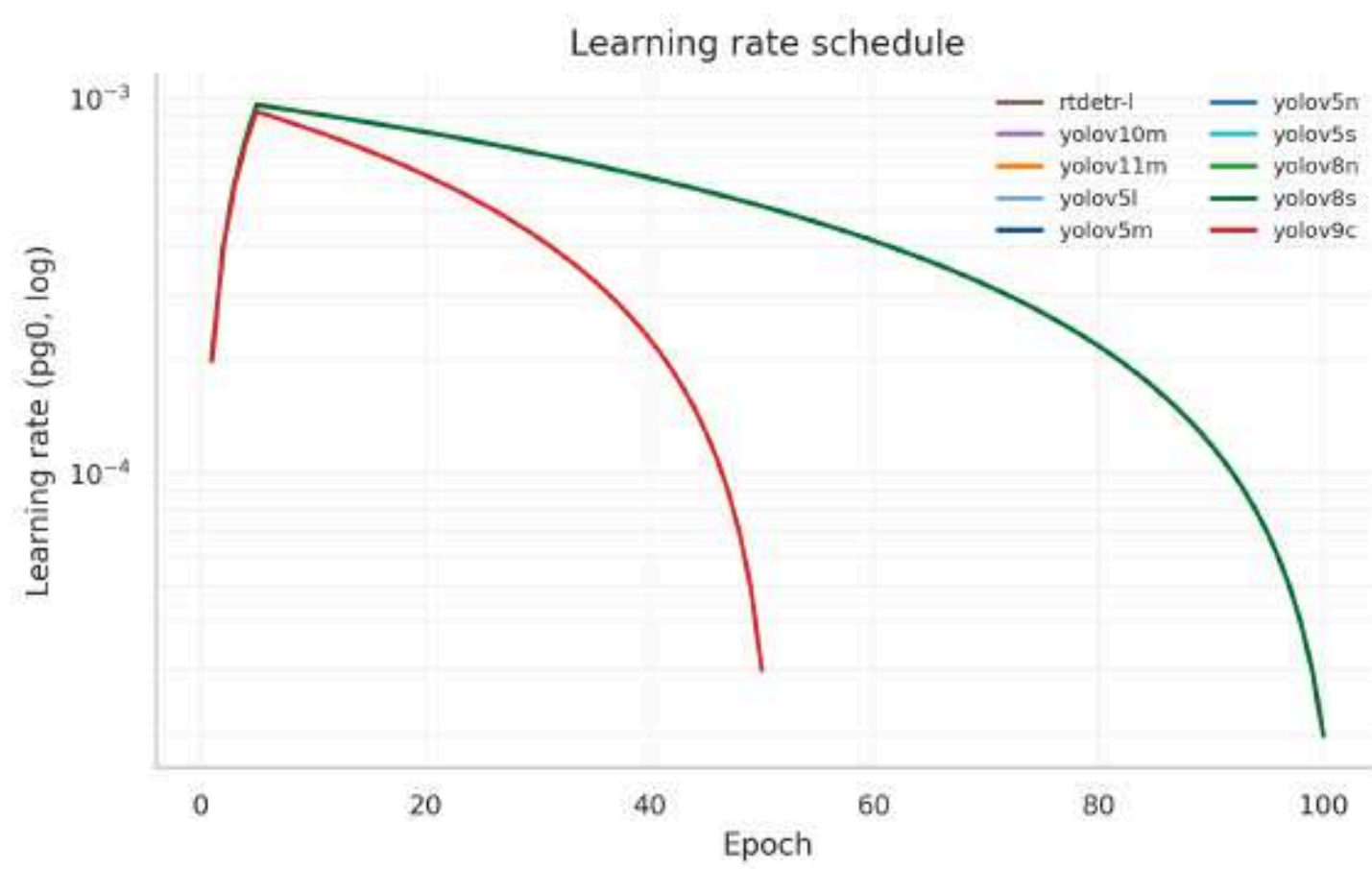

Figure 9: Learning rate schedule across training. All models use cosine annealing from $10^{-3}$ to $10^{-5}$.

Figure 11 compares the best mAP achieved by each model during training at both IoU thresholds. These training-time peaks closely match the final benchmark numbers in Table 5, confirming that the models did not overfit and that the early stopping criterion preserved the best checkpoint reliably.

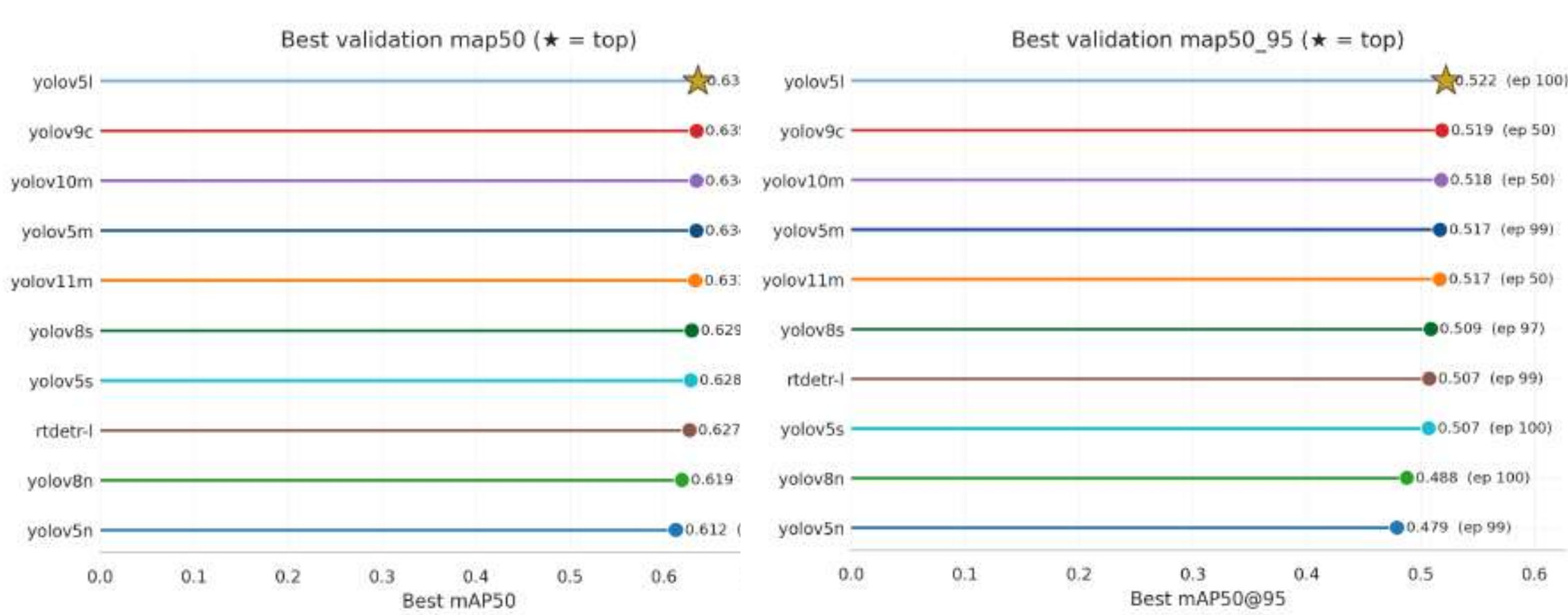

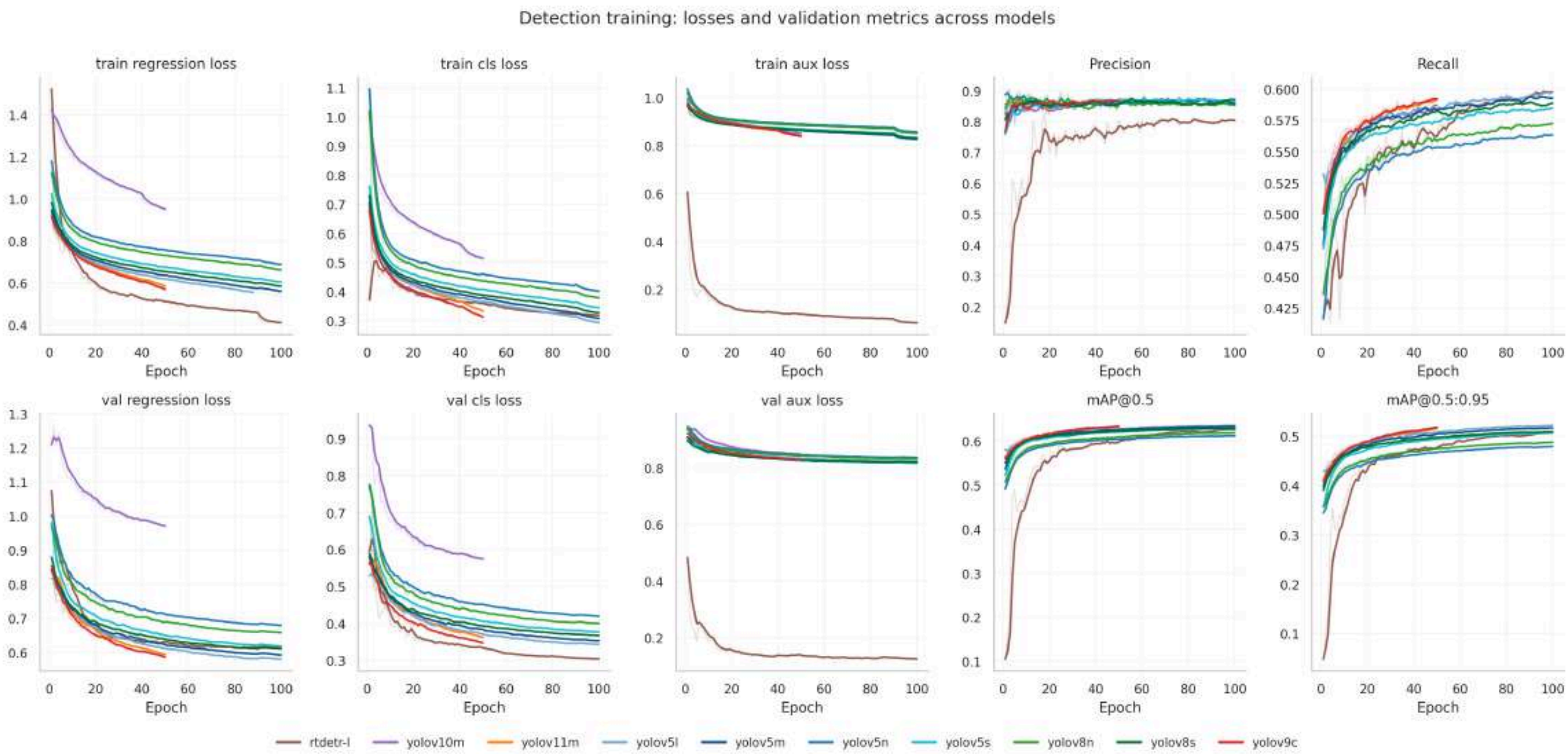

Figure 10: Comprehensive training curves for all ten detector variants, showing loss, precision, recall, and mAP evolution across epochs.

Figure 11: Best mAP@50 (left) and mAP@50-95 (right) achieved during training for each model.

### 6.5 Qualitative Results

Figure 12 along with Figure 13 show representative validation predictions from the best model (YOLOv5l). The detector correctly localizes vehicles in dense urban scenes, including partially occluded vehicles and vehicles at varying scales. Road regions are detected but with less precise boundaries, consistent with the quantitative per-class analysis above.

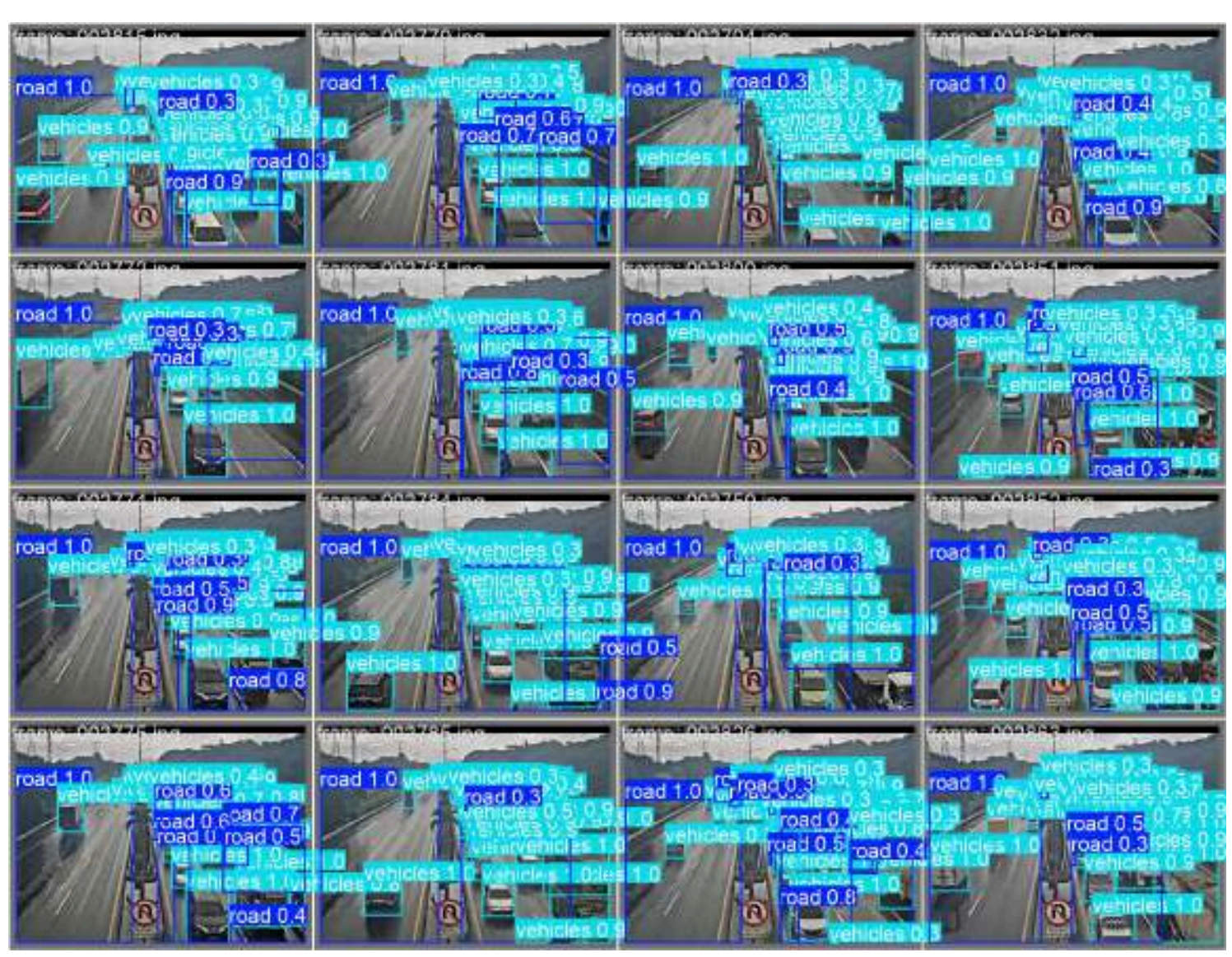

Figure 12: Validation predictions from YOLOv5l on representative test frames. Bounding boxes show detected vehicles and road regions.

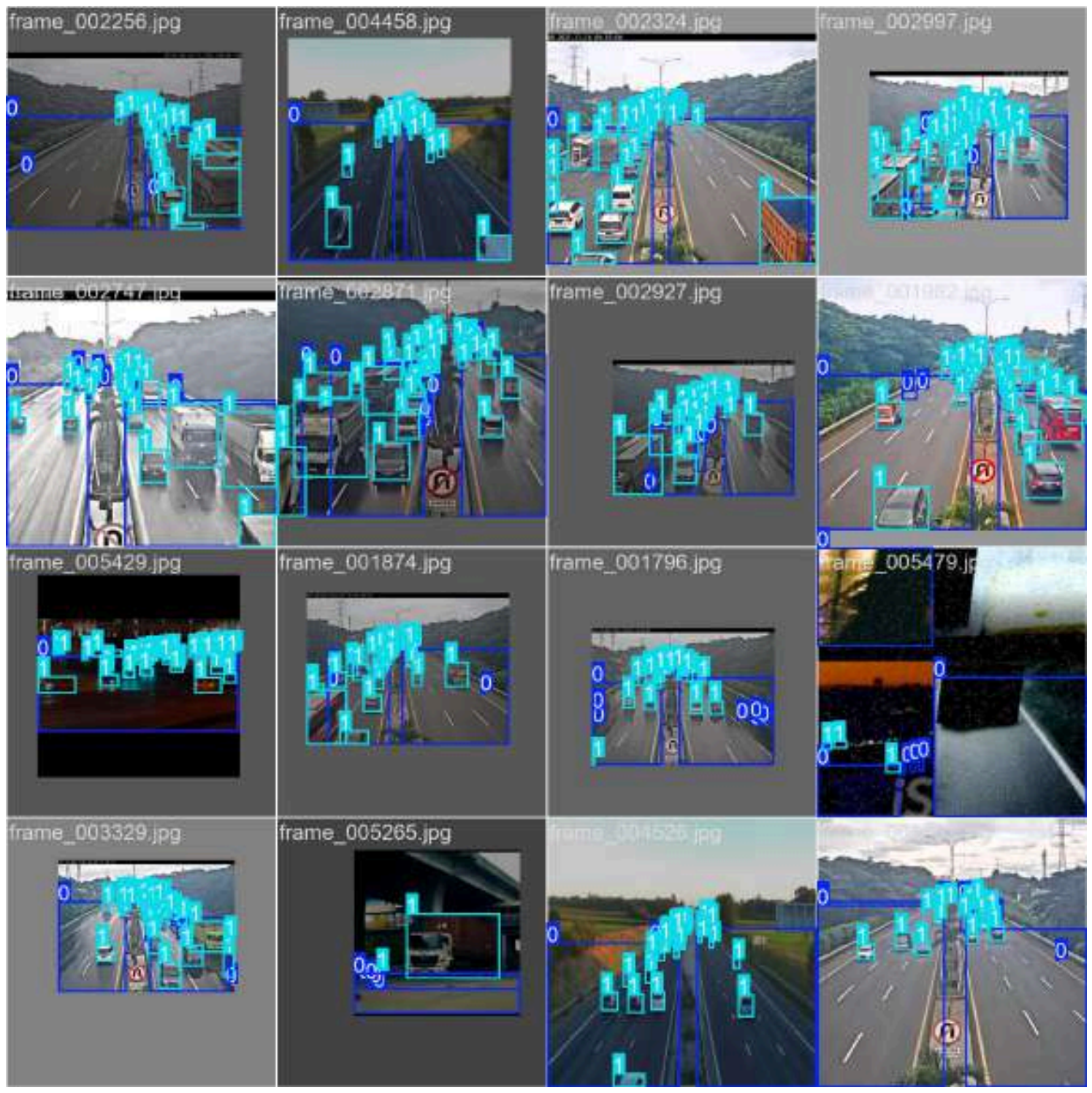

Figure 13: Validation predictions from YOLOv5l on representative training frames. Bounding boxes show detected vehicles and road regions.

Figure 14 and Figure 15 provide direct visual comparisons of detection accuracy. Figure 14 shows mAP at both IoU thresholds across all ten models; the narrow spread is immediately apparent, reinforcing the model-agnostic nature of the pipeline. Figure 15 consolidates all key metrics, precision, recall, mAP@50, mAP@50-95, F1, and FPS, into a single multi-panel view, making it straightforward to compare the full performance profile of each model at a glance.

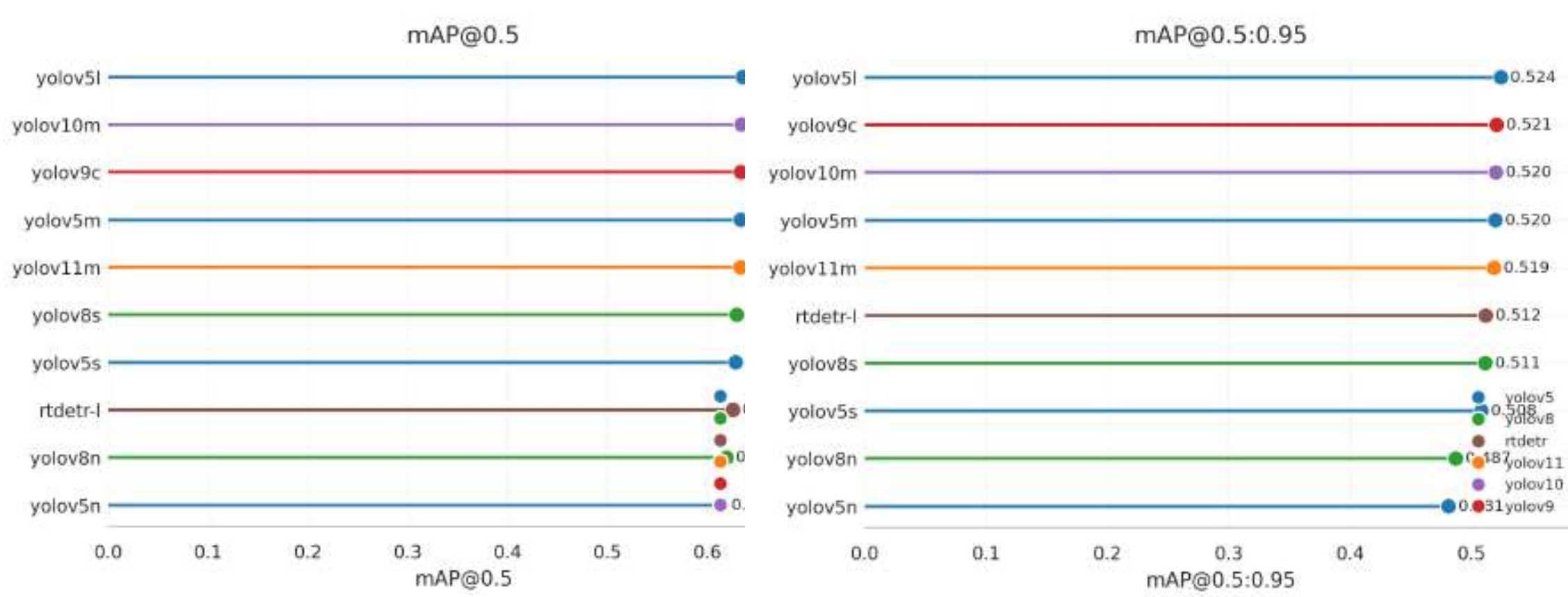

Figure 14: mAP@50 (left) and mAP@50-95 (right) comparison across all detector variants.

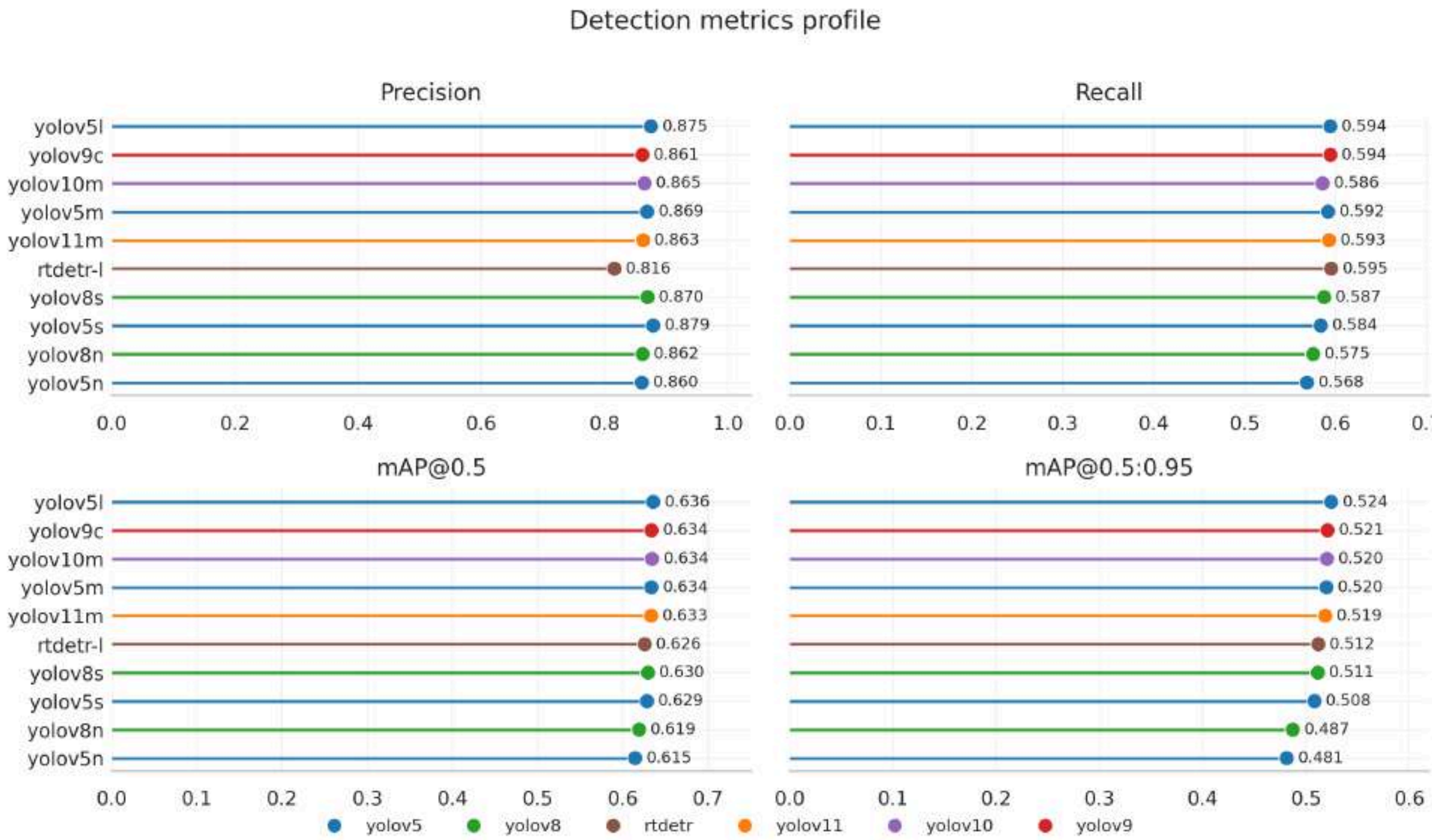

Figure 15: Multi-panel benchmark comparison grid showing precision, recall, mAP@50, mAP@50-95, F1, and FPS across all models.

### 6.6 Temporal Estimator Results

We now evaluate Stage 6 of the pipeline: the temporal estimators trained on the cleaned YOLOv5l telemetry stream. Table 6 reports overall metrics for all three architectures on the held-out evaluation stream (29,243 windows, 5-step horizon).

Table 6: Overall estimator performance on the held-out stream (29,243 windows; YOLOv5l telemetry). Metrics are in original (un-normalized) feature units. Models ranked by $R^2$; bold marks the best value per column.

| Model | MSE | MAE | RMSE | $R^2$ | RMSE T+1 | RMSE T+3 | RMSE T+5 |
|---|---|---|---|---|---|---|---|
| GRU | 2 875 695 | 440.92 | **1695.79** | **0.858** | **1565.83** | 1742.07 | 1756.84 |
| LSTM | 2 915 943 | 439.08 | 1707.61 | 0.856 | 1645.60 | 1753.46 | 1728.72 |
| TCN | 3 228 263 | 471.08 | 1796.74 | 0.840 | 1776.64 | 1858.69 | 1743.93 |

The GRU achieves the best overall performance ($R^2 = 0.858$, RMSE = 1695.79), narrowly ahead of the LSTM (0.856) and the TCN (0.840). The three architectures fall within roughly 6 % of one another on RMSE, so no single recipe is decisive on this task; the gated recurrent units edge out the dilated convolutions by virtue of their slightly tighter T+1 forecast (GRU: 1565.83 vs. TCN: 1776.64). This ordering inverts the qualitative result obtained in earlier exploratory runs that used a smaller upstream detector (YOLOv8n): there, the higher upstream noise level appeared to favor the TCN's larger receptive field, whereas the cleaner YOLOv5l telemetry reduces the marginal benefit of the convolutional architecture and lets the recurrent gates shine on short windows. The practical takeaway is that the choice of estimator architecture is, like the choice of detector, secondary to the spatial logic layer that produces the telemetry in the first place.

Figure 16 shows RMSE as a function of forecast step. The GRU maintains the lowest T+1 error and remains competitive through T+5; the LSTM tracks closely, while the TCN starts higher at T+1 and converges with the others by T+5. Importantly, all three curves rise from T+1 to mid-horizon rather than remaining flat, which confirms that the estimators learn genuine short-window temporal structure (a copy-last-frame baseline would produce a flat or erratic horizon profile).

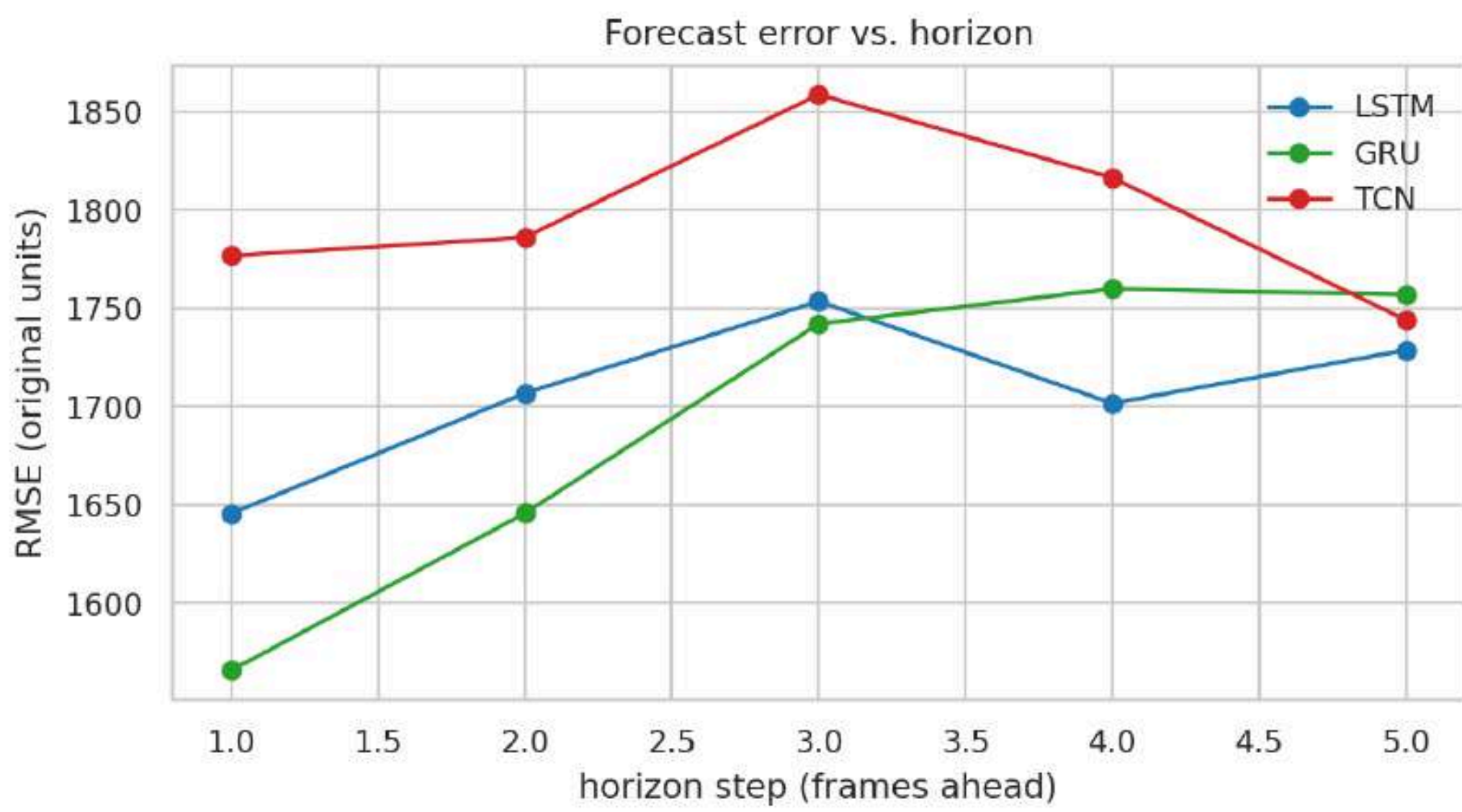

Figure 16: RMSE as a function of forecast horizon step (T+1 through T+5) for all three estimators on YOLOv5l telemetry. The GRU leads at T+1; all three converge by T+5.

Figure 17 presents the per-feature RMSE on a log scale, making the multi-order-of-magnitude range across features visible. As before, direction_variance dominates the absolute error budget by several orders of magnitude, which means the overall $R^2$ of $\approx$ 0.86 is driven primarily by the model's ability to track that one high-magnitude feature. We therefore report per-feature behaviour qualitatively through this figure rather than as a single aggregate, since aggregate $R^2$ alone overstates how well the estimator performs on the operationally important low-magnitude features (vehicle_count, occupancy_ratio, congestion_index).

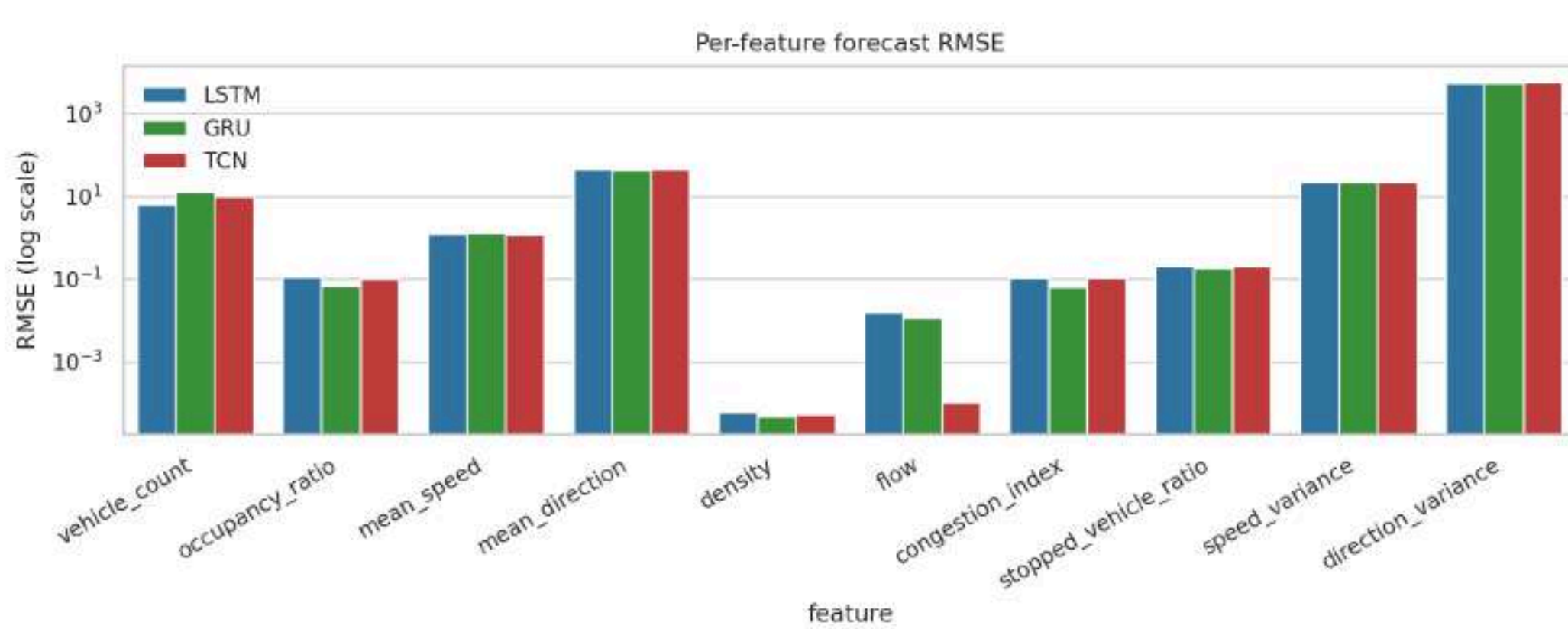

Figure 17: Per-feature RMSE across all three estimators (log scale) on the YOLOv5l evaluation stream. The multi-order-of-magnitude range across features explains why overall $R^2$ is dominated by high-magnitude features.

Figure 18 shows predicted-versus-actual time series for the three operationally relevant features. On vehicle_count, all three estimators track the underlying mean and capture broad rises and falls but miss most of the high-frequency transients caused by individual vehicles entering or leaving the frame. On occupancy_ratio and congestion_index, the GRU produces visibly tighter tracking than the TCN, consistent with the aggregate metrics in Table 6.

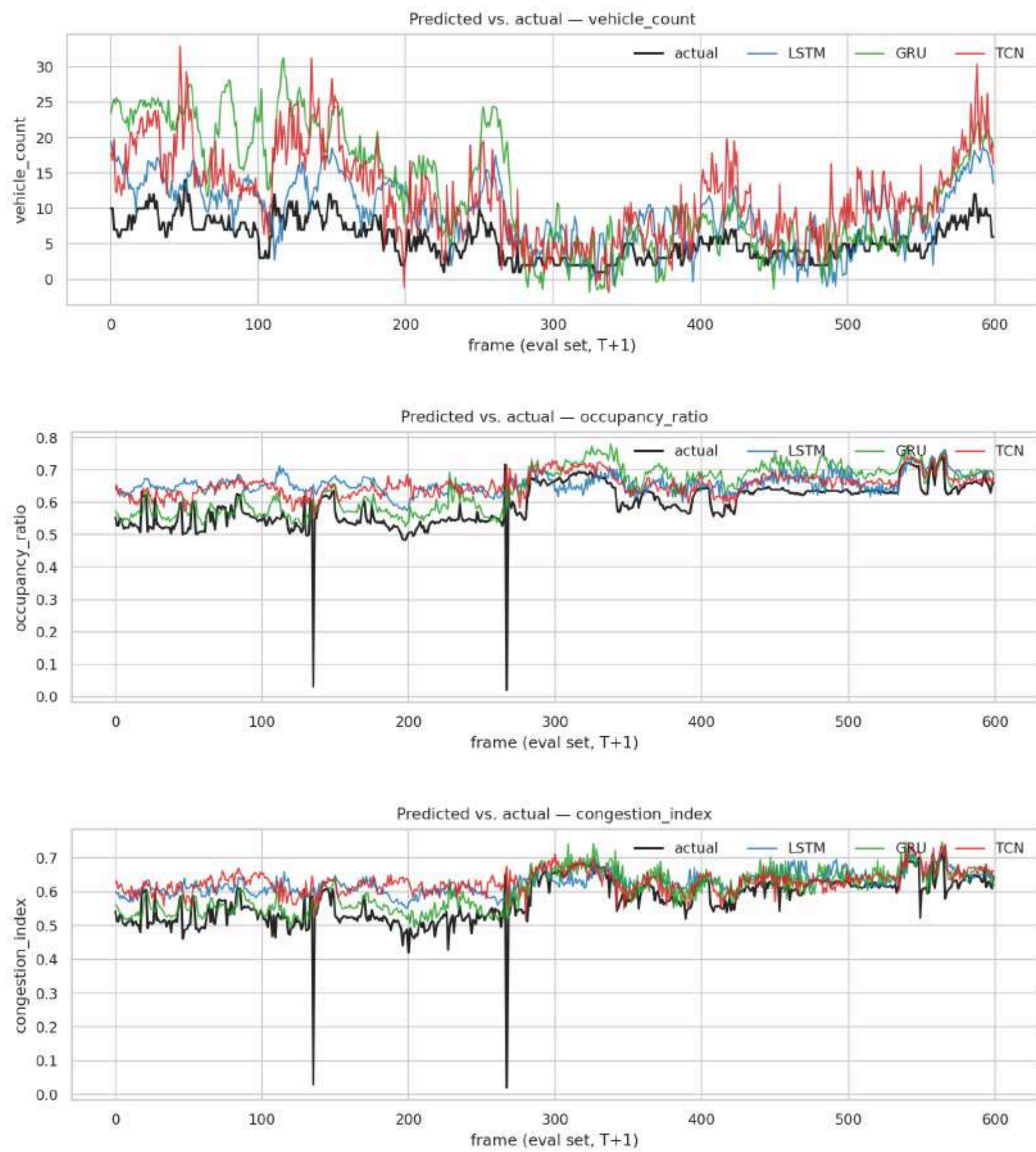

Figure 18: Predicted vs. actual time series for vehicle_count (top), occupancy_ratio (middle), and congestion_index (bottom) on the held-out YOLOv5l evaluation stream. All three estimators are overlaid against the ground truth.

Figure 19 shows the prediction-versus-truth scatter for the three operationally relevant features. The vehicle_count panel concentrates near the origin, reflecting an evaluation stream dominated by low-count frames with sporadic high-count bursts the models struggle to anticipate. The occupancy_ratio and congestion_index panels show tighter clustering along the diagonal, consistent with their narrower dynamic range, but all three estimators visibly under-predict at the upper tail of each feature.

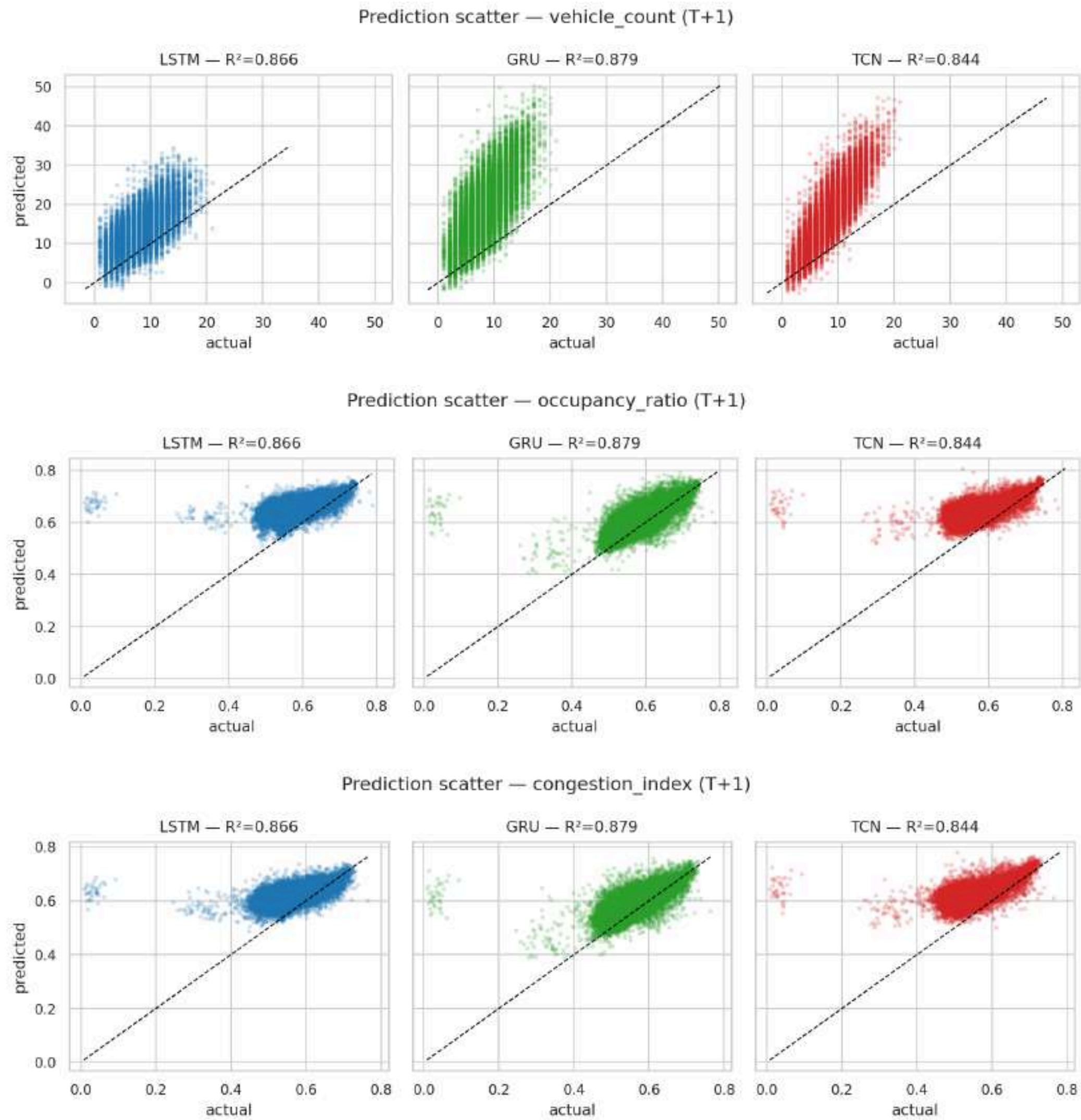

Figure 19: Scatter of predicted vs. actual values for vehicle_count (top), occupancy_ratio (middle), and congestion_index (bottom). The diagonal denotes perfect prediction.

### 6.7 Live Pipeline Telemetry

The offline metrics above evaluate each estimator on a single static window of held-out telemetry. To validate that the same estimator behaviour holds when the full six-stage pipeline runs end-to-end, we deploy the system live on a third Caltrans stream with the YOLOv5l detector and each estimator in turn. The pipeline writes per-frame predictions for horizons T+1 through T+5 alongside the realized feature values, and we compare them after the fact.

Figure 20 shows the rolling T+1 prediction RMSE (50-frame window) for the three operationally relevant features across all three estimators. The error envelopes are stable rather than growing — there is no evidence of drift or accumulation under the multi-threaded inference path — and the LSTM/GRU/TCN traces remain close to one another throughout, mirroring the narrow margins reported in Table 6. Brief excursions in the rolling RMSE coincide with traffic-state transitions in the underlying stream rather than with anything specific to one architecture, indicating that the estimators degrade and recover synchronously when the upstream telemetry shifts.

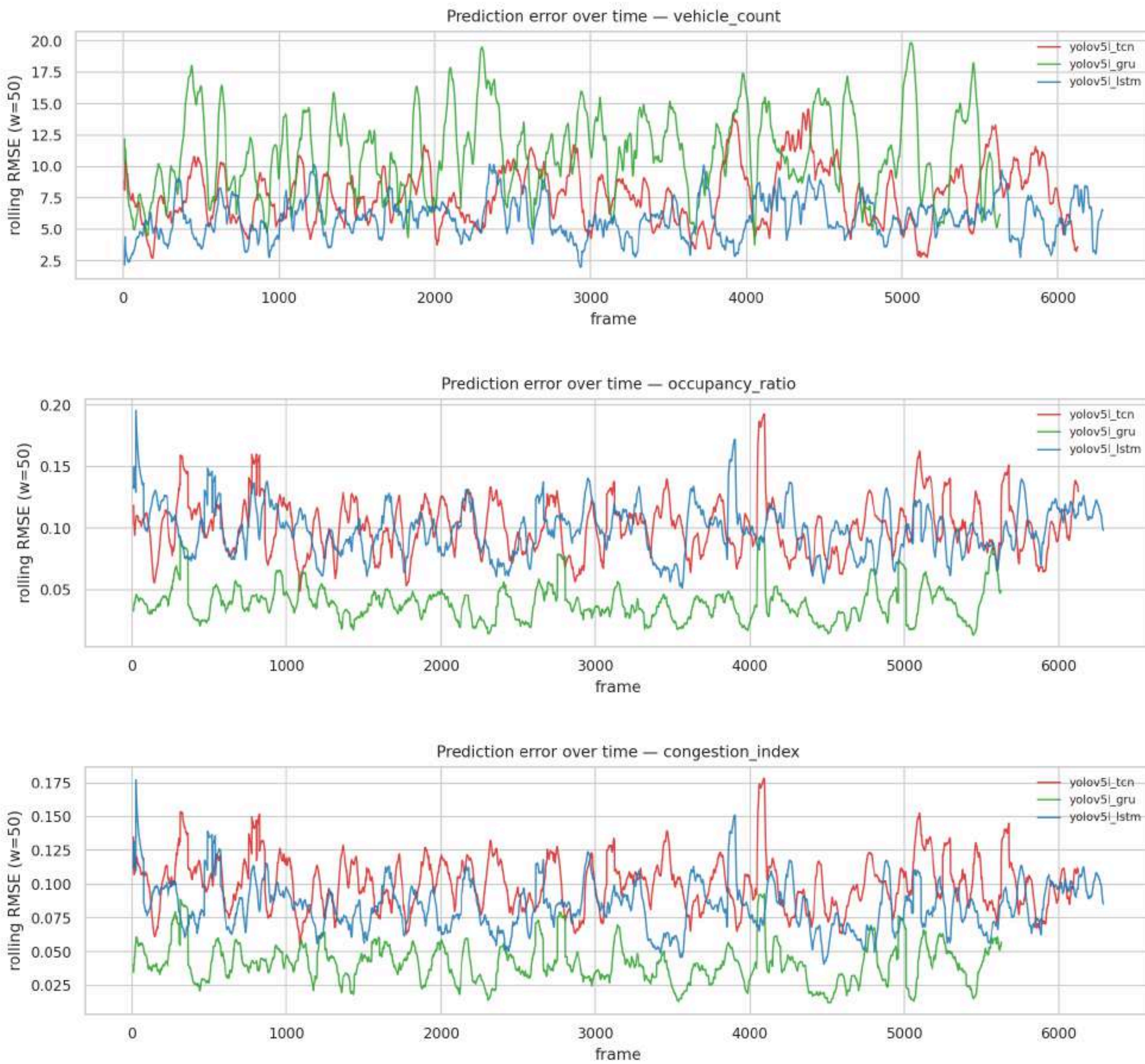

Figure 20: Rolling T+1 prediction RMSE (window = 50 frames) over a single live YOLOv5l pipeline run, for vehicle_count (top), occupancy_ratio (middle), and congestion_index (bottom). The three estimator traces remain close, with excursions reflecting underlying traffic-state transitions rather than architecture-specific failure modes.

Figure 21 visualises how the multi-step forecast envelopes track the realized feature values frame-by-frame. For each estimator we plot the actual line and shade the region between actual and the T+1..T+5 predictions; a tighter band indicates better calibration across the full horizon. The GRU and LSTM bands hug the actual line most closely on occupancy_ratio and congestion_index, while all three estimators widen on vehicle_count whenever the count spikes — consistent with the per-feature analysis in Figure 18.

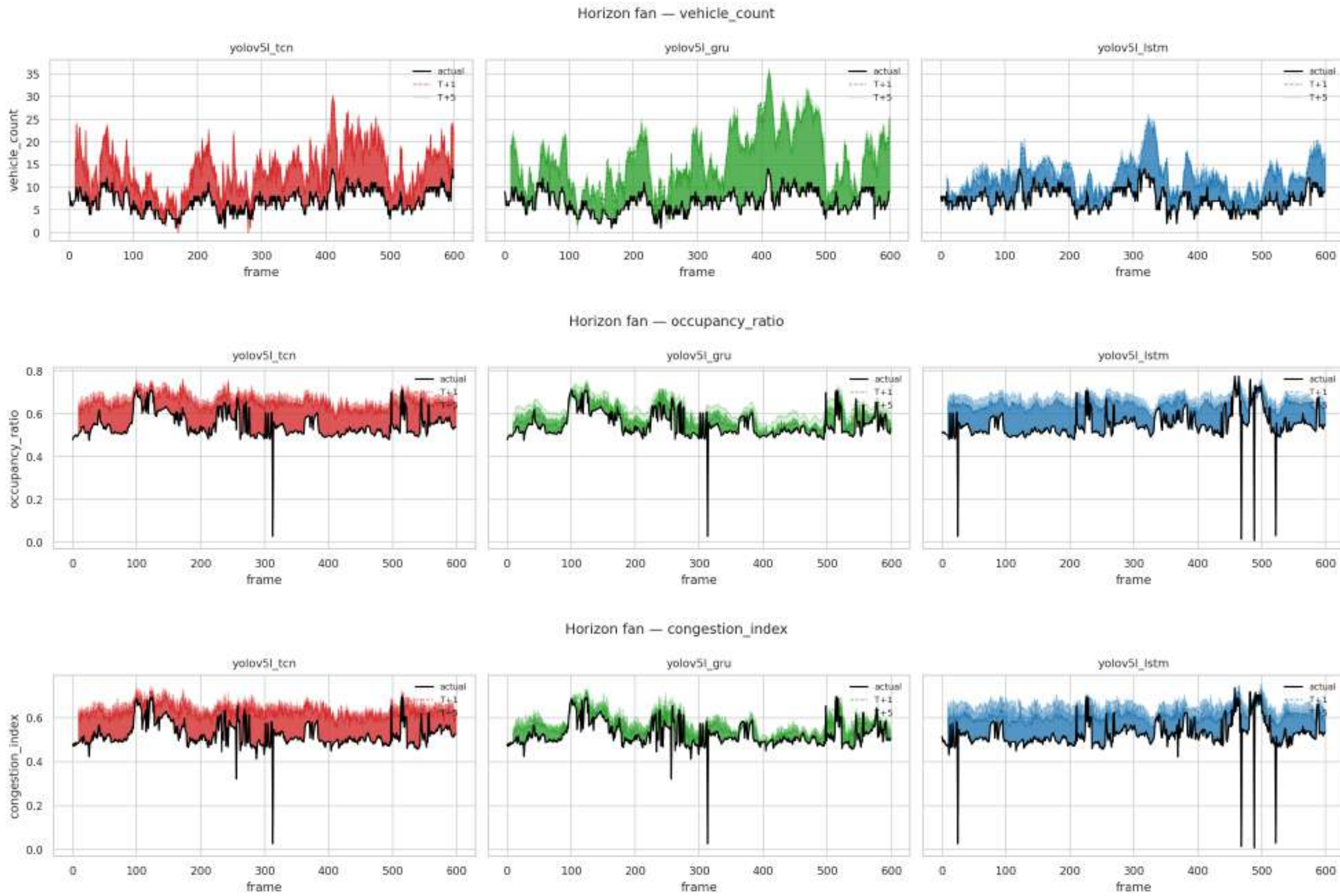

Figure 21: Live-pipeline horizon fans for the three operationally relevant features. Each panel column corresponds to one estimator (LSTM/GRU/TCN); the black line is the realized feature value and the shaded region spans the T+1..T+5 predictions. Tighter bands indicate better forecast calibration across the full horizon.

### 6.8 Discussion

The detection and estimation results together support four conclusions about the pipeline.

First, the pipeline is genuinely model-agnostic on the detection side: swapping detector families changes throughput substantially but affects accuracy only marginally, reducing the deployment decision to a hardware-budget question rather than an architecture-selection problem. Second, the Pareto frontier provides actionable guidance: YOLOv8n or YOLOv5n for edge devices at 250+ FPS, YOLOv8s or YOLOv5s for balanced deployments at 180+ FPS, and YOLOv5l or YOLOv9c for accuracy-critical installations at 56–64 FPS. Third, RT-DETR, despite its architectural novelty, does not offer a favorable trade-off on this dataset, likely because the transformer's global attention mechanism provides less benefit on the relatively uniform traffic scenes in our data than it does on the diverse object categories of COCO.

Fourth, on the estimation side, the three architectures (LSTM, GRU, TCN) deliver near-equivalent performance, with the GRU narrowly leading on the YOLOv5l telemetry stream ($R^2 = 0.858$, RMSE = 1695.79; GRU vs. TCN within 6 % on RMSE). The horizon-step decomposition shows the GRU's advantage is concentrated at T+1, with all three models converging by T+5. This near-tie suggests that on cleaner upstream telemetry the gated recurrent units are sufficient to capture the short-window dynamics, and the TCN's larger receptive field offers no decisive benefit over a 10-frame lookback. The architectural choice therefore mirrors the detection-side conclusion: it is a deployment-budget decision (GRUs are the smallest and fastest of the three) rather than an accuracy-driven one.

The per-feature analysis reveals an important nuance: the overall $R^2$ of $\approx 0.86$ is dominated by direction_variance, which has large absolute values. This metric should not be interpreted as indicating strong predictive performance across all features. The high-variance, operationally important features — vehicle_count, occupancy_ratio, and congestion_index — remain challenging because their frame-to-frame dynamics are driven by discrete detection events that a 10-frame lookback cannot reliably anticipate.

The live-pipeline run (Section 6.7) provides a final consistency check. The rolling T+1 RMSE traces remain stable across the run, with no architecture-specific drift, and the three estimator bands stay close throughout. This confirms that the offline metrics generalize to the multi-threaded end-to-end inference path: queue dynamics, frame drops, and the FPS controller do not introduce additional estimator-side error.

### 6.8.1 Limitations

Several limitations qualify these results. The estimator is trained on a single camera stream and evaluated on one held-out stream (offline) plus one live stream (end-to-end), which is sufficient to confirm internal consistency but limits claims about cross-camera generalization. The estimator forecasts the detector's own feature stream, not ground-truth traffic density; the reported $R^2$ therefore measures self-consistency of the pipeline rather than absolute traffic-density accuracy. The high-magnitude direction_variance feature dominates the aggregate $R^2$, which obscures persistently weak per-feature performance on vehicle_count and occupancy_ratio, the two features most directly relevant to traffic management. The 10-frame lookback window may be too short to capture longer traffic cycles such as signal phases or platoon arrivals. Finally, no road mask was available for the evaluation stream, so occupancy, density, and congestion metrics include off-road pixels, inflating noise in those features.

## 7 Conclusion

This paper presents a two-stage pipeline for real-time traffic density estimation that explicitly addresses the interface between visual perception and temporal prediction. The spatial logic layer, which filters detections by comparing bounding boxes against a precomputed road mask, is the key component that transforms noisy detector output into a clean telemetry stream suitable for downstream forecasting. Our evaluation across ten models from six detector families confirms that the pipeline is model-agnostic and that the choice of detector reduces to a throughput-versus-accuracy trade-off governed by the deployment hardware budget.

The temporal estimator evaluation, using YOLOv5l telemetry, confirms that all three architectures (LSTM, GRU, TCN) learn meaningful short-window temporal structure, with the GRU narrowly achieving the best overall performance ($R^2 = 0.858$, RMSE = 1695.79) ahead of the LSTM (0.856) and the TCN (0.840). The architectures fall within 6 % of one another on RMSE, which mirrors the detection-side finding that comparable models converge to comparable accuracy and that the spatial logic layer, not the choice of estimator architecture, is the distinguishing component. A live end-to-end run on a previously-unseen camera stream further confirms that these offline metrics carry over to the multi-threaded inference path: rolling prediction error remains stable and the three estimator traces stay close throughout.

The Pareto frontier we report provides concrete deployment guidance: lightweight models (YOLOv5n, YOLOv8n) sustain 250+ FPS for embedded edge nodes, mid-range models (YOLOv8s, YOLOv5s) offer a balanced operating point at 180+ FPS, and larger models (YOLOv5l, YOLOv9c) maximize accuracy at 56–64 FPS for GPU-accelerated installations.

Future work targets improvements at each pipeline stage independently. At the detection stage, optimized vision transformer backbones, attention-based feature pyramids, and knowledge distillation from larger to smaller models can push the accuracy-throughput frontier further. At the spatial logic stage, dynamic road masking that adapts to construction zones and temporary lane closures would improve robustness in real-world deployments. At the feature extraction stage, incorporating environmental context — weather conditions, ambient lighting, and time-of-day — as additional input dimensions should help the estimator disambiguate traffic-induced variance from environmental noise. At the estimation stage, transformer-based sequence models, architecture search over the TCN topology, and physics-informed hybrid losses that embed macroscopic flow constraints may improve per-feature $R^2$ on the currently weak features. Beyond stage-specific improvements, training on multi-camera, multi-city telemetry streams with longer lookback windows and extended forecast horizons is the most direct path to improving generalization and per-feature accuracy. Finally, replacing bounding-box occupancy with instance segmentation masks would yield finer-grained density estimates when hardware permits.

## Bibliography

[1] B. Dwyer and J. Nelson, "Roboflow." 2022.

[2] G. Jocher, "YOLOv5 by Ultralytics." [Online]. Available: https://github.com/ultralytics/yolov5

[3] G. Jocher, A. Chaurasia, and J. Qiu, "Ultralytics YOLOv8." [Online]. Available: https://github.com/ultralytics/ultralytics

[4] C.-Y. Wang, I.-H. Yeh, and H.-Y. M. Liao, "YOLOv9: Learning What You Want to Learn Using Programmable Gradient Information." 2024.

[5] A. Wang *et al.*, "YOLOv10: Real-Time End-to-End Object Detection." 2024.

[6] G. Jocher and J. Qiu, "Ultralytics YOLO11." [Online]. Available: https://github.com/ultralytics/ultralytics

[7] W. Lv *et al.*, "DETRs Beat YOLOs on Real-time Object Detection." 2024.

[8] M. N. Sadik, T. Hossain, and F. Sayeed, "Real-Time Detection and Analysis of Vehicles and Pedestrians using Deep Learning." 2024.

[9] J. Nubert and others, "Traffic Density Estimation using a Convolutional Neural Network." 2018.

[10] K. G. Zoysa and S. R. Munasinghe, "Vision-Based Incoming Traffic Estimator Using Deep Neural Network on General Purpose Embedded Hardware." 2023.

[11] M. Pereira, A. Lang, and B. Kulcsár, "Short-Term Traffic Prediction Using Physics-Aware Neural Networks." 2021.

[12] D. Wilkman and others, "Online Traffic Density Estimation using Physics-Informed Neural Networks." 2025.

[13] B. Wang and others, "Training Physics-Informed Neural Networks via Multi-Task Optimization for Traffic Density Prediction," in *2023 International Joint Conference on Neural Networks (IJCNN)*, 2023.

[14] N. Aharon, R. Orfaig, and B.-Z. Bobrovsky, "BoT-SORT: Robust Associations Multi-Pedestrian Tracking." 2022.

[15] Y. Zhang *et al.*, "ByteTrack: Multi-Object Tracking by Associating Every Detection Box," in *European Conference on Computer Vision (ECCV)*, 2022.

[16] S. Hochreiter and J. Schmidhuber, "Long Short-Term Memory," *Neural Computation*, vol. 9, no. 8, pp. 1735–1780, 1997.

[17] K. Cho *et al.*, "Learning Phrase Representations using RNN Encoder–Decoder for Statistical Machine Translation," in *Proceedings of the 2014 Conference on Empirical Methods in Natural Language Processing (EMNLP)*, 2014, pp. 1724–1734.

[18] S. Bai, J. Z. Kolter, and V. Koltun, "An Empirical Evaluation of Generic Convolutional and Recurrent Networks for Sequence Modeling." 2018.

# 8 Appendix: Per-Model Detailed Results

This appendix presents individual training curves, normalized confusion matrices, precision-recall curves, and F1-confidence curves for each of the ten detector variants. These complement the aggregated views in the main body (Figure 10, Figure 15) by revealing model-specific convergence behaviour.

## 8.1 YOLOv5n

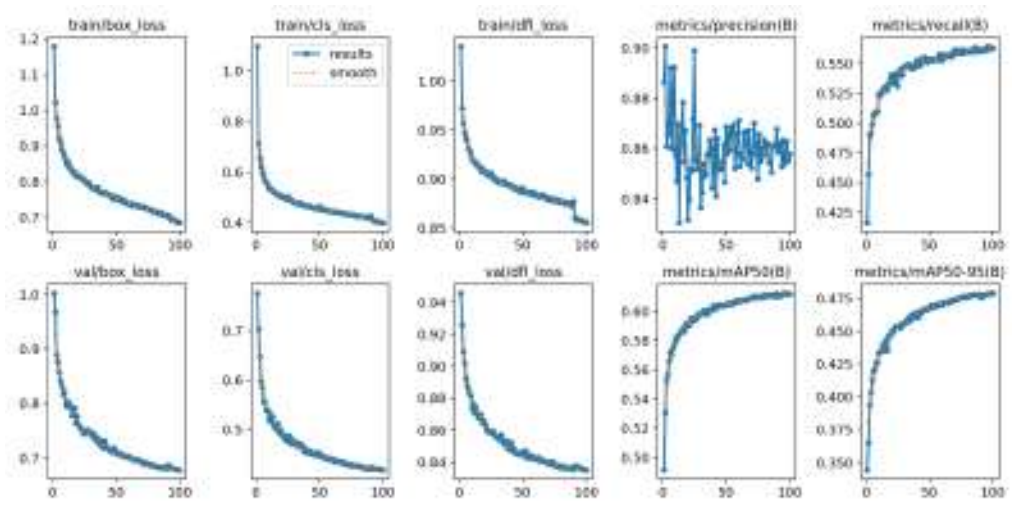

Figure 22: Training curves.

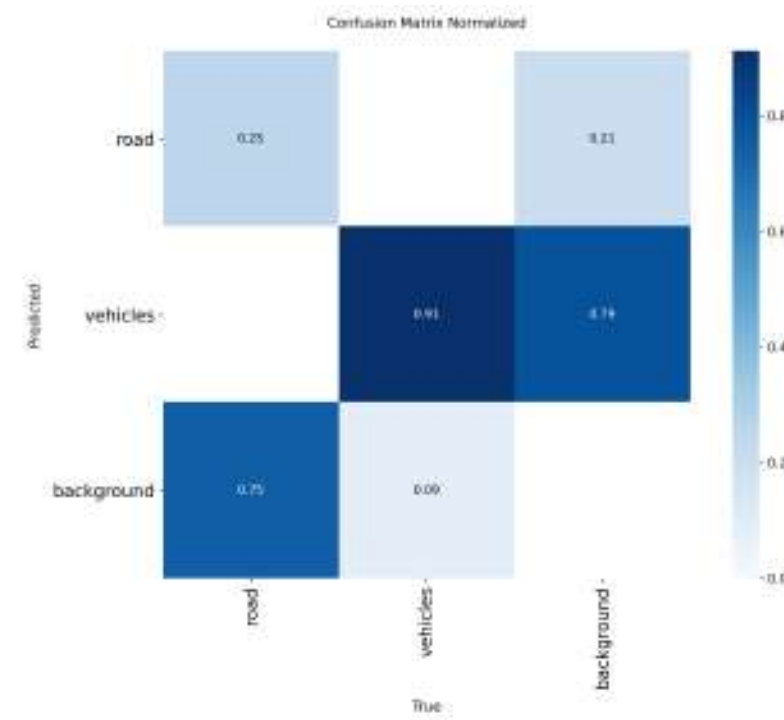

Figure 23: Confusion matrix.

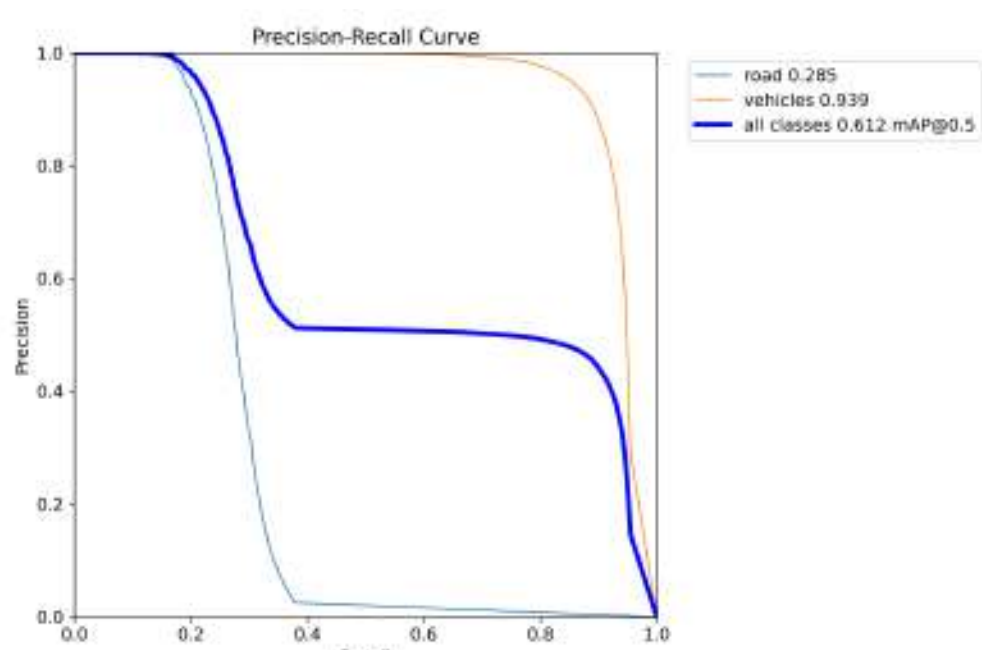

Figure 24: PR curve.

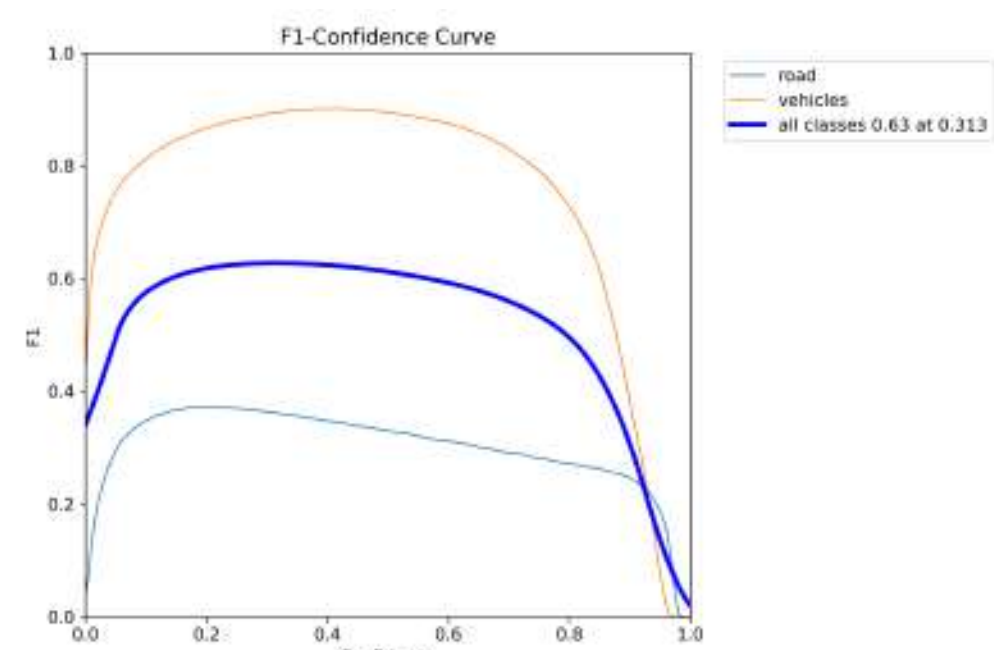

Figure 25: F1 curve.

## 8.2 YOLOv5s

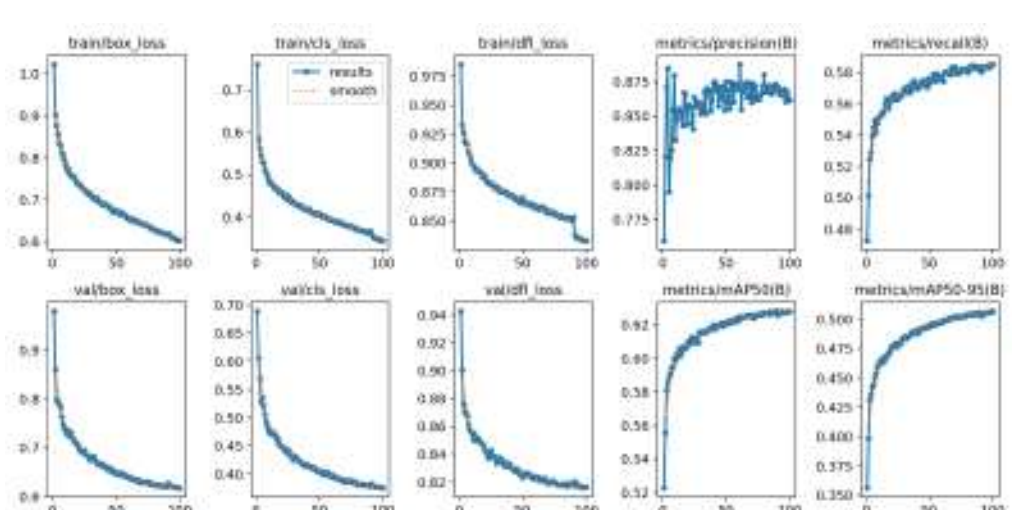

Figure 26: Training curves.

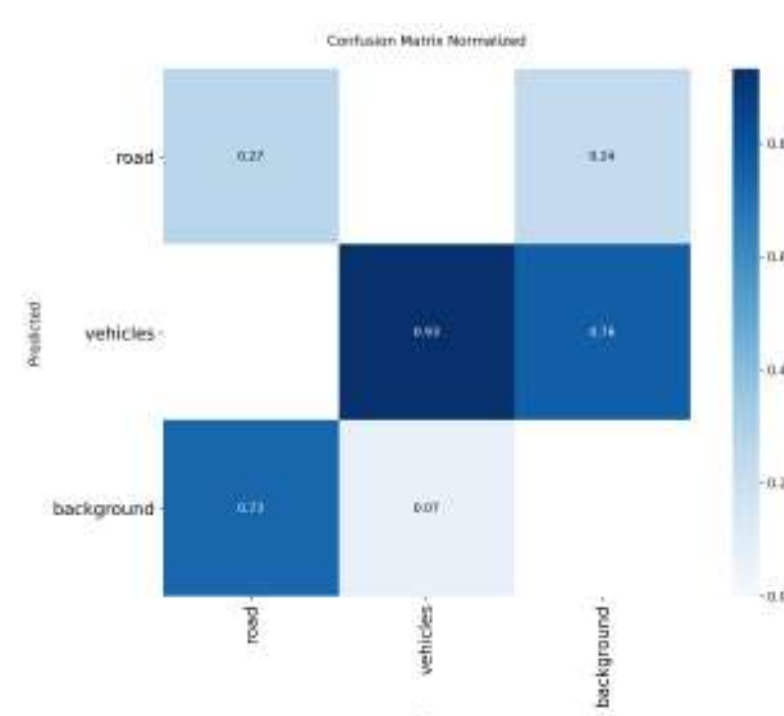

Figure 27: Confusion matrix.

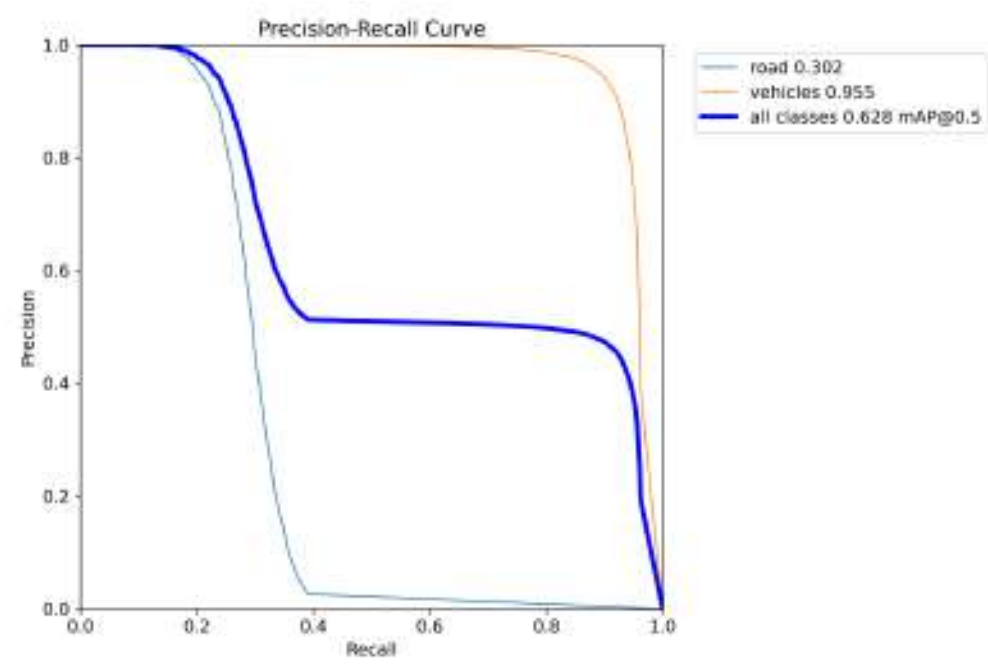

Figure 28: PR curve.

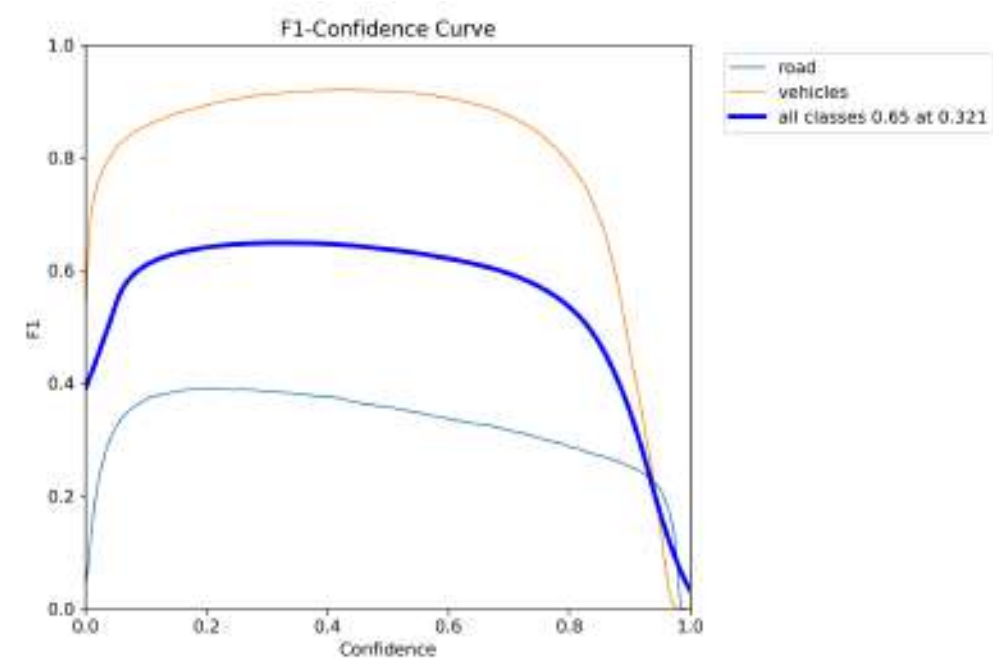

Figure 29: F1 curve.

## 8.3 YOLOv5m

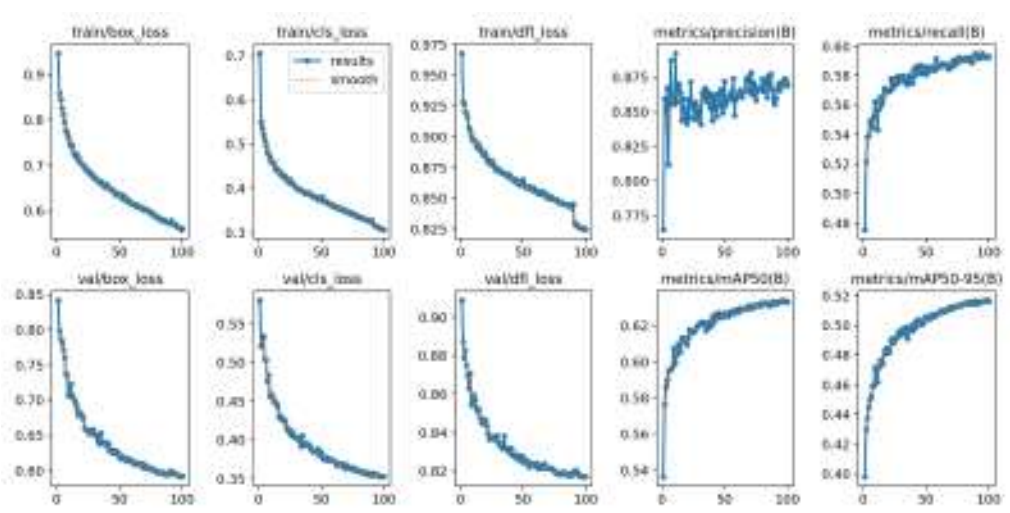

Figure 30: Training curves.

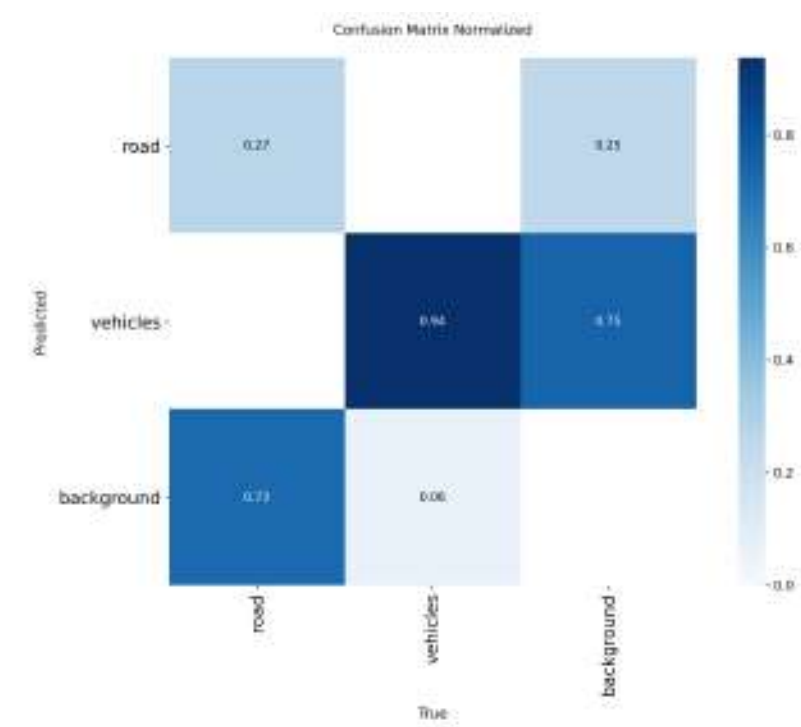

Figure 31: Confusion matrix.

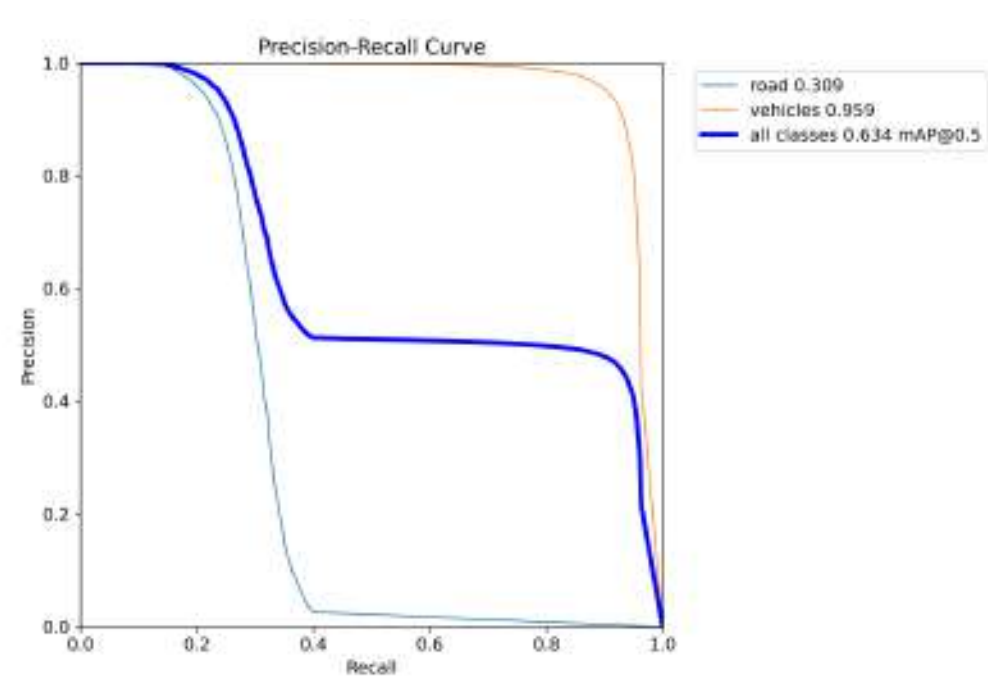

Figure 32: PR curve.

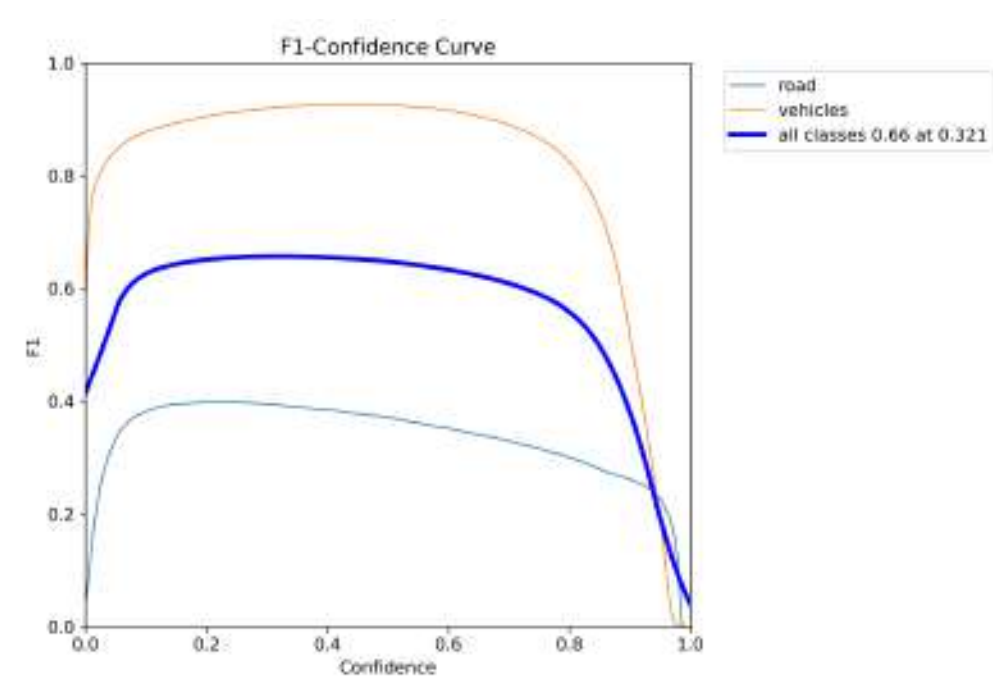

Figure 33: F1 curve.

## 8.4 YOLOv5l

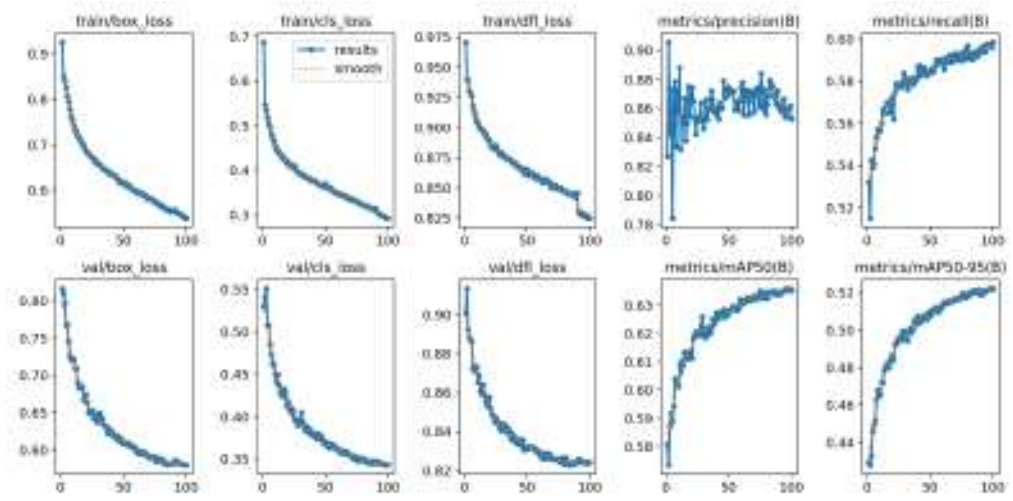

Figure 34: Training curves.

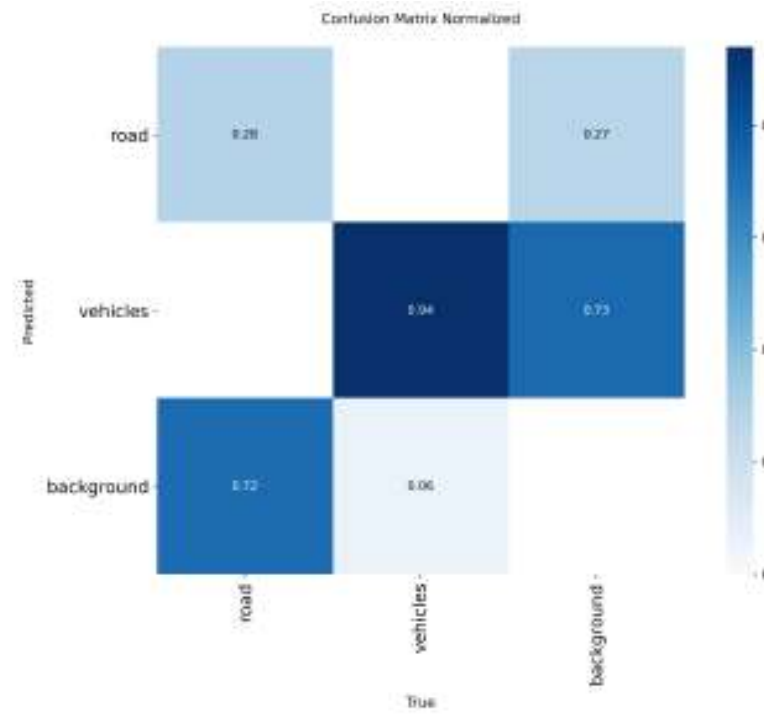

Figure 35: Confusion matrix.

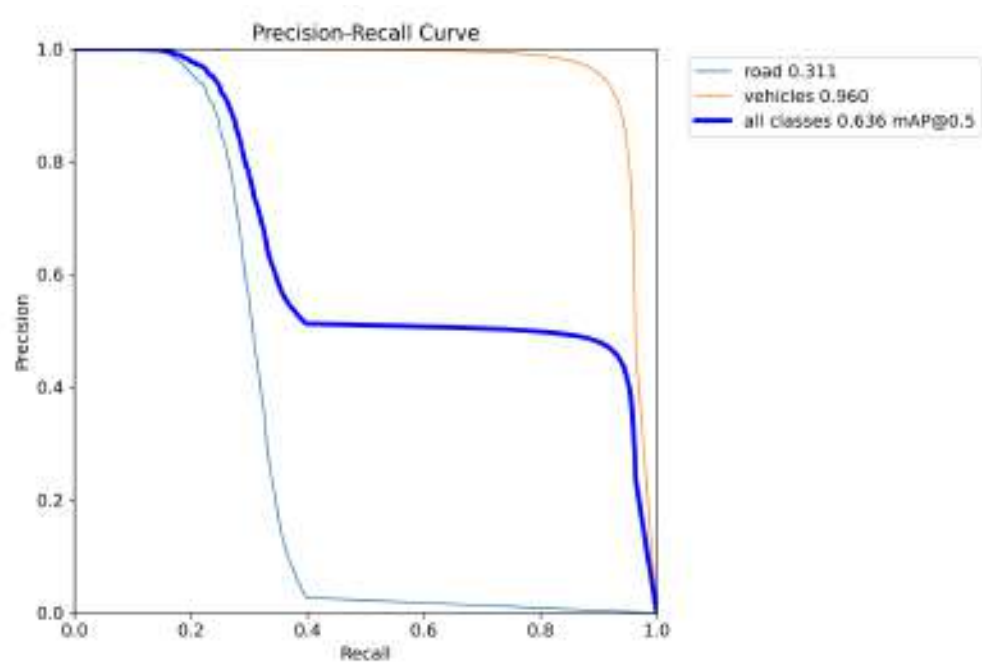

Figure 36: PR curve.

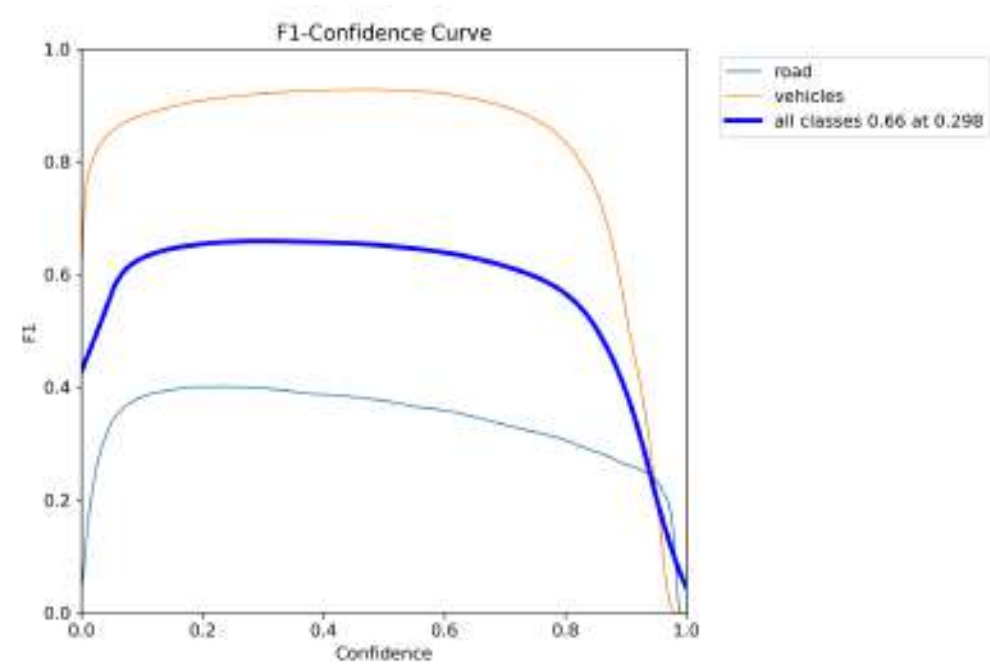

Figure 37: F1 curve.

## 8.5 YOLOv8n

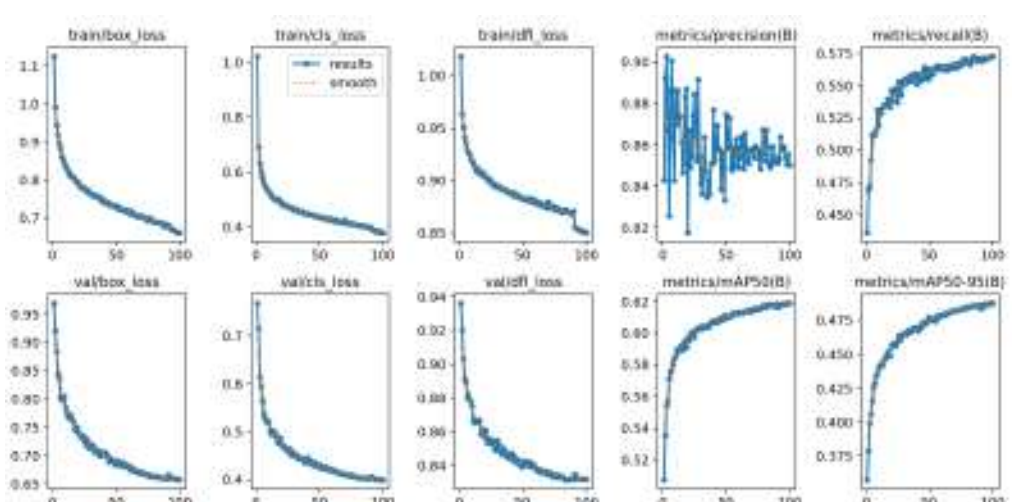

Figure 38: Training curves.

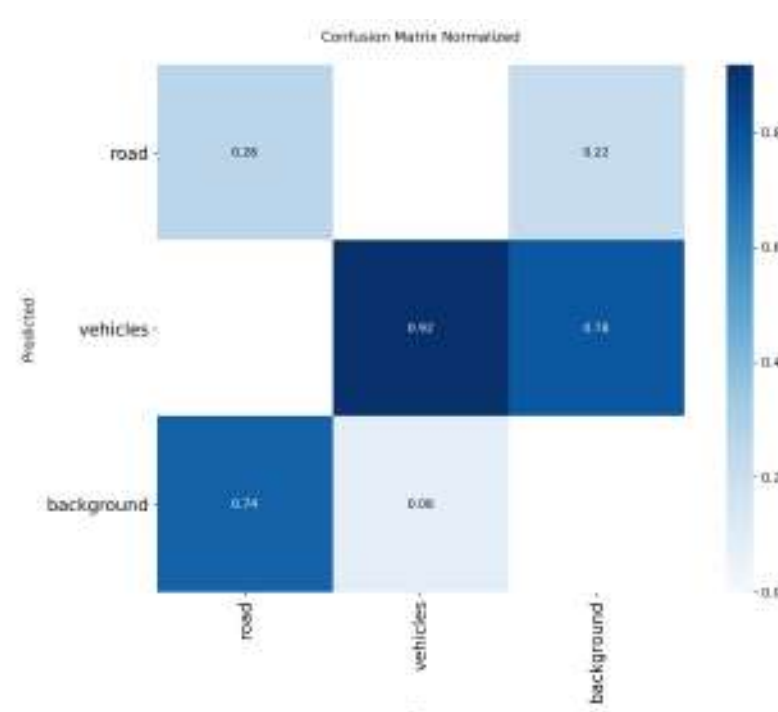

Figure 39: Confusion matrix.

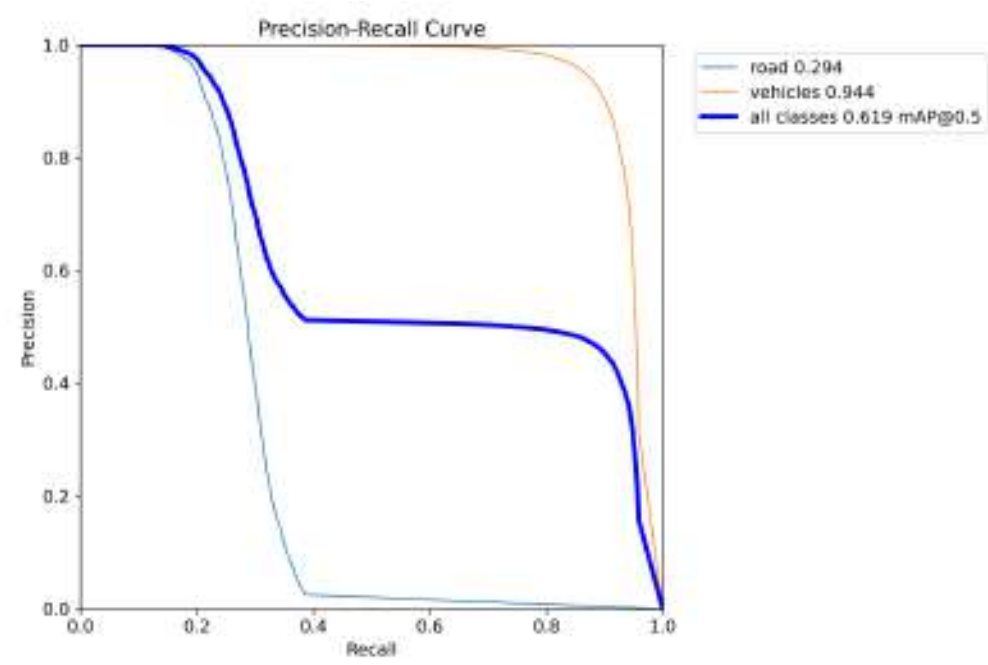

Figure 40: PR curve.

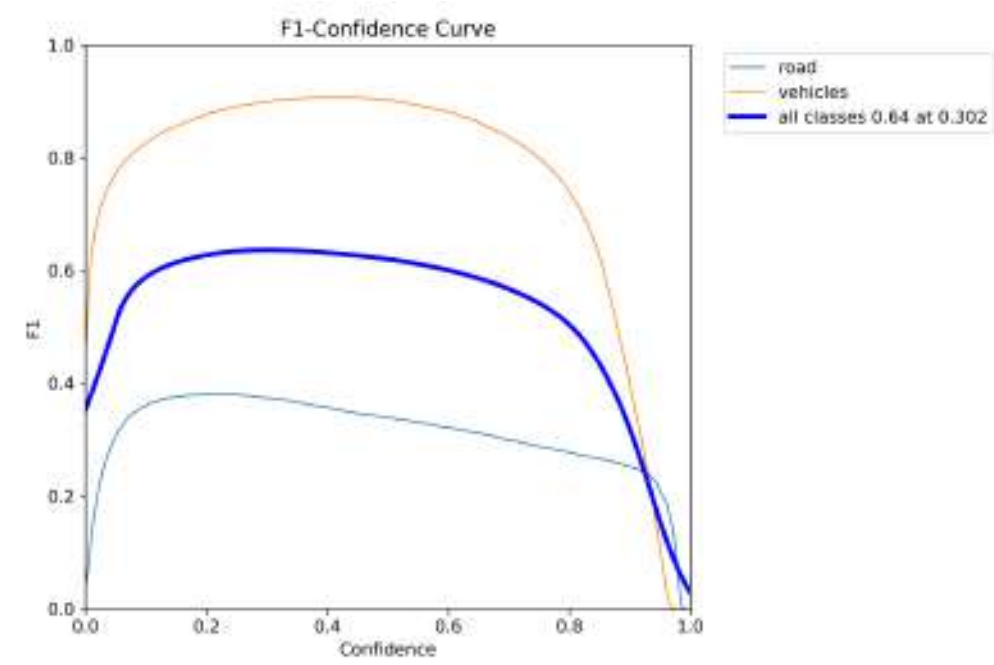

Figure 41: F1 curve.

## 8.6 YOLOv8s

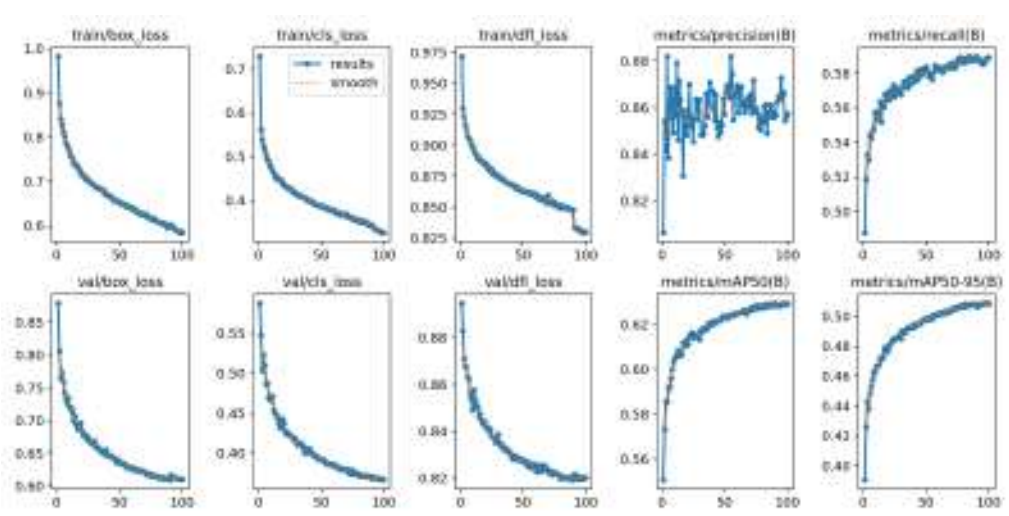

Figure 42: Training curves.

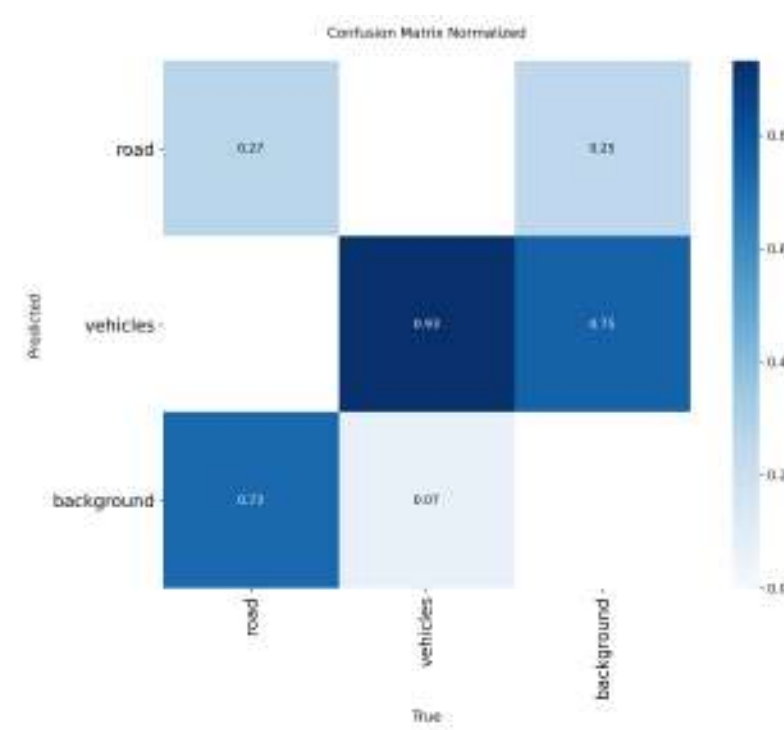

Figure 43: Confusion matrix.

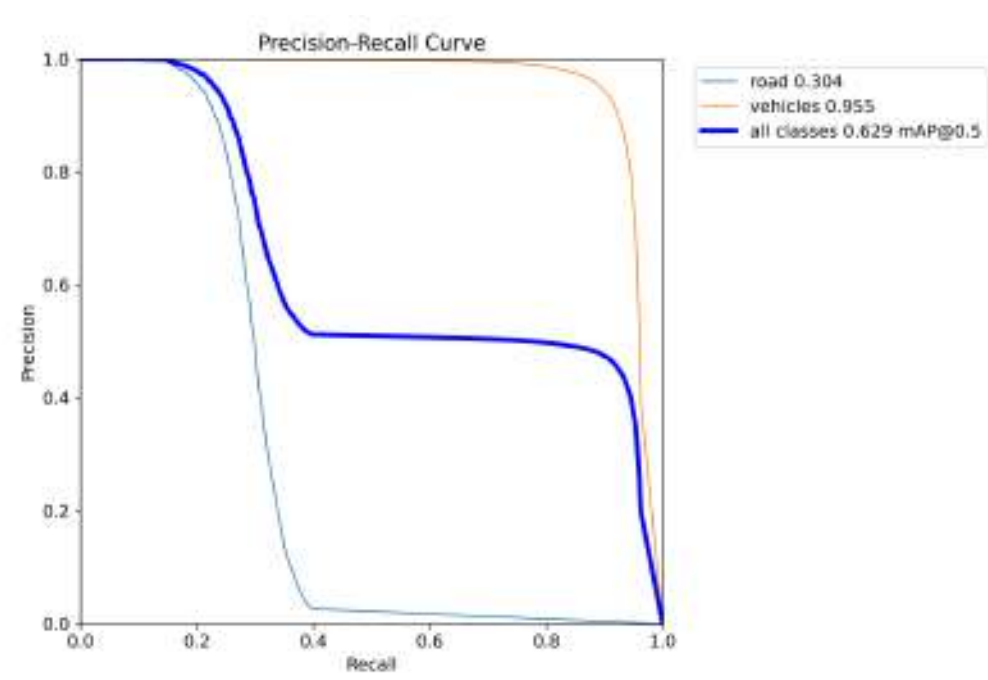

Figure 44: PR curve.

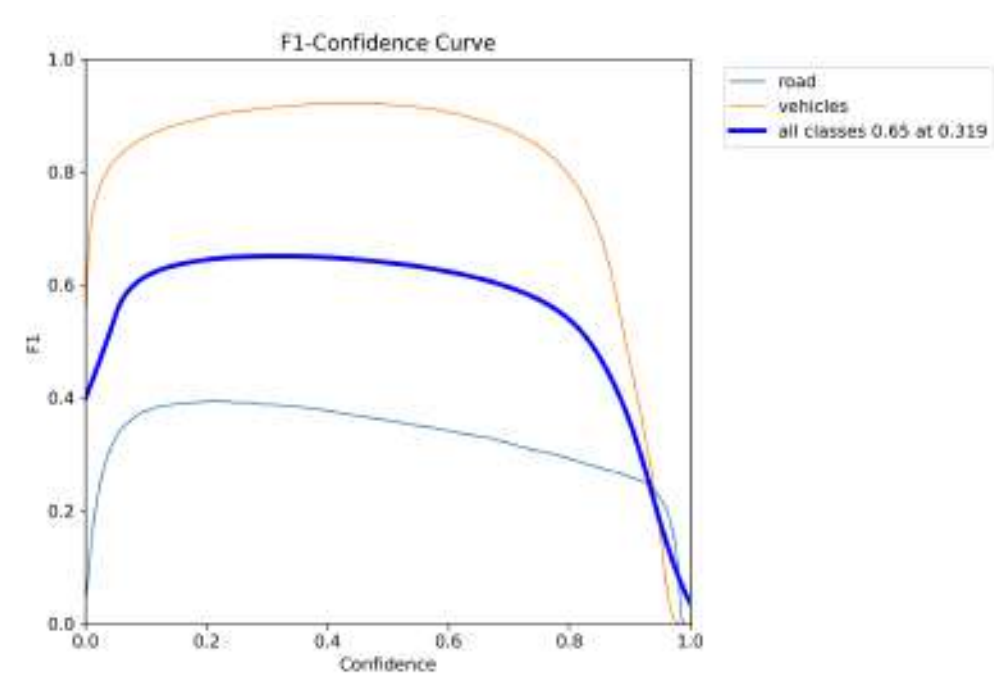

Figure 45: F1 curve.

## 8.7 YOLOv9c

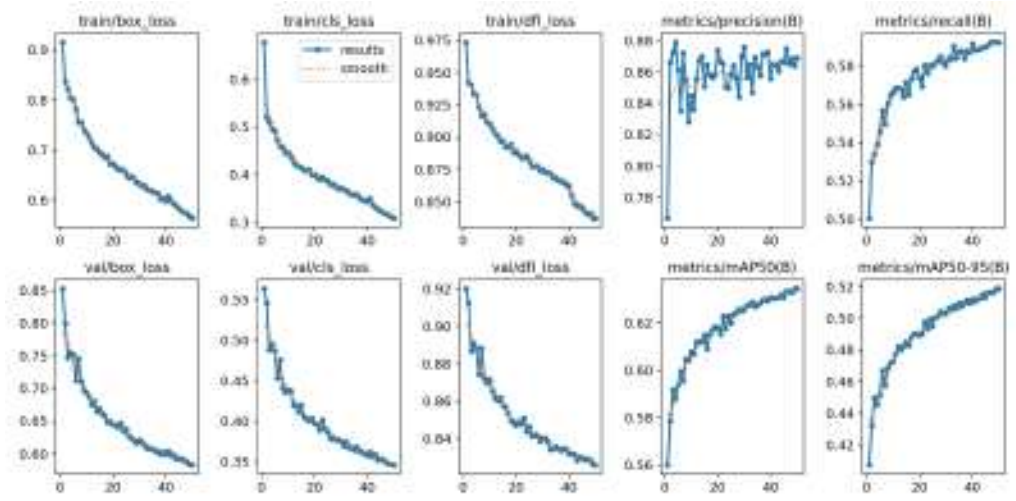

Figure 46: Training curves.

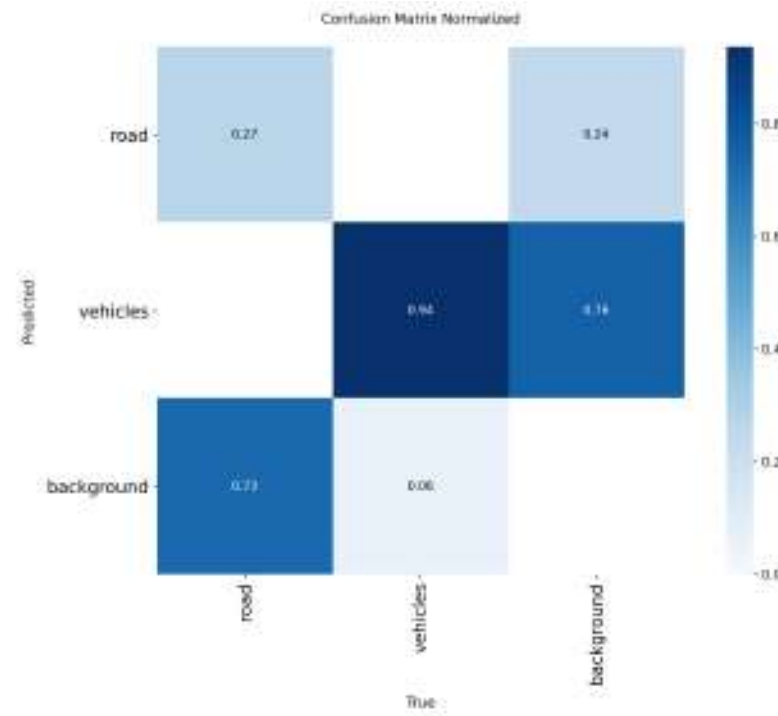

Figure 47: Confusion matrix.

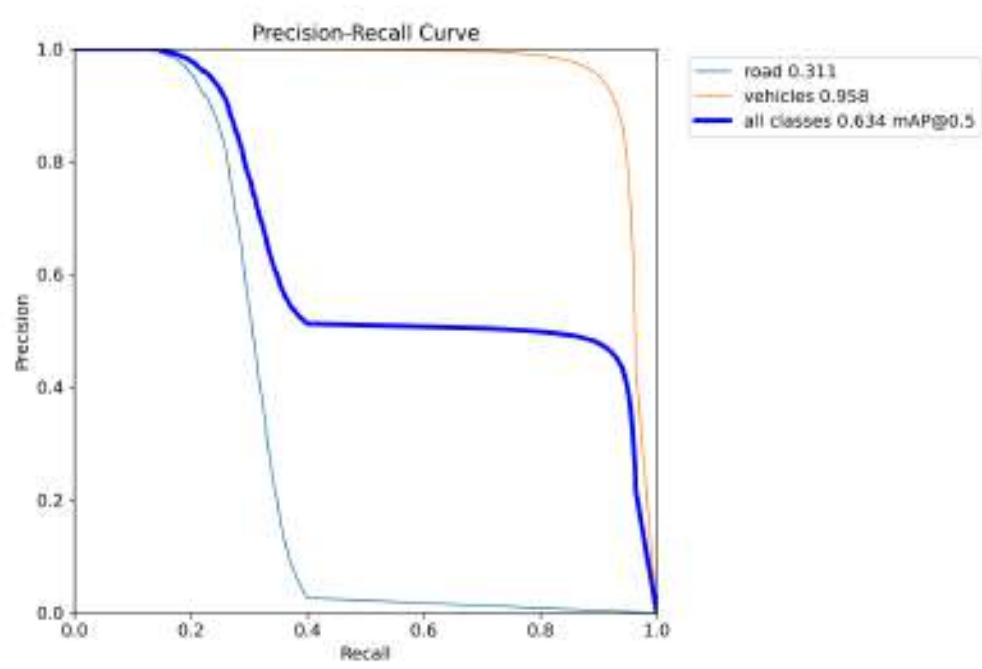

Figure 48: PR curve.

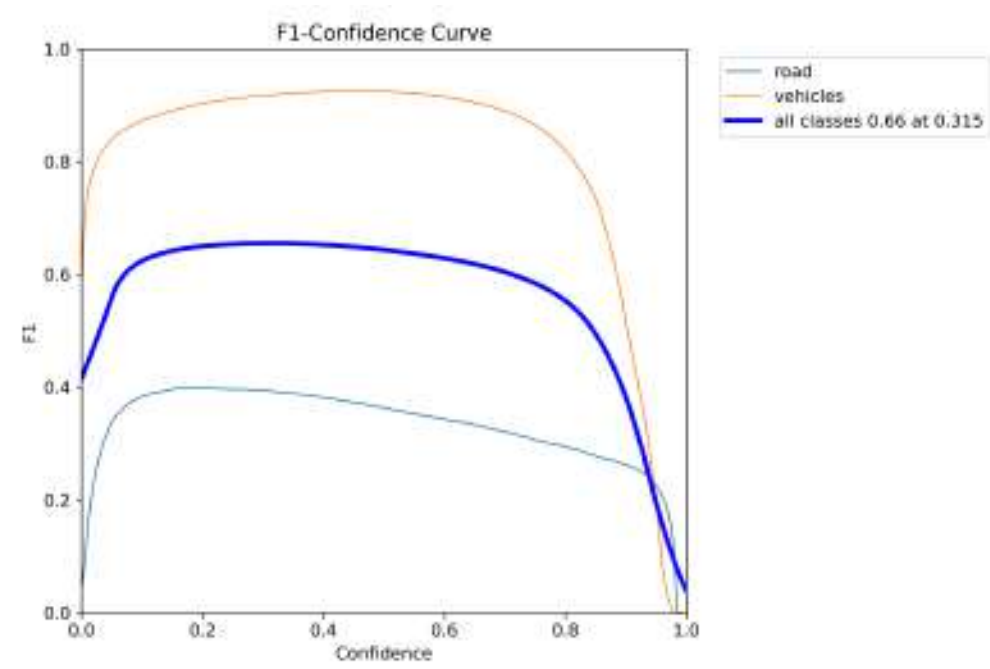

Figure 49: F1 curve.

## 8.8 YOLOv10m

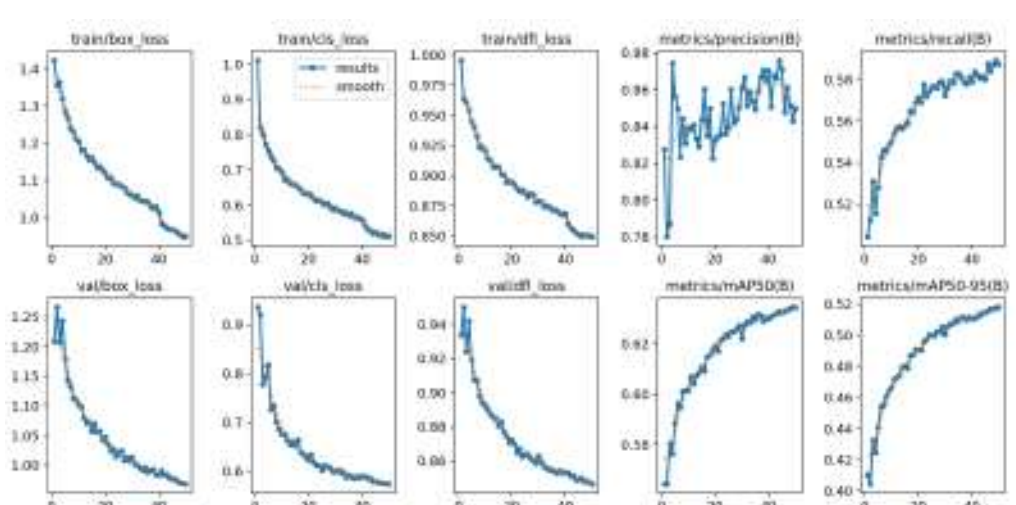

Figure 50: Training curves.

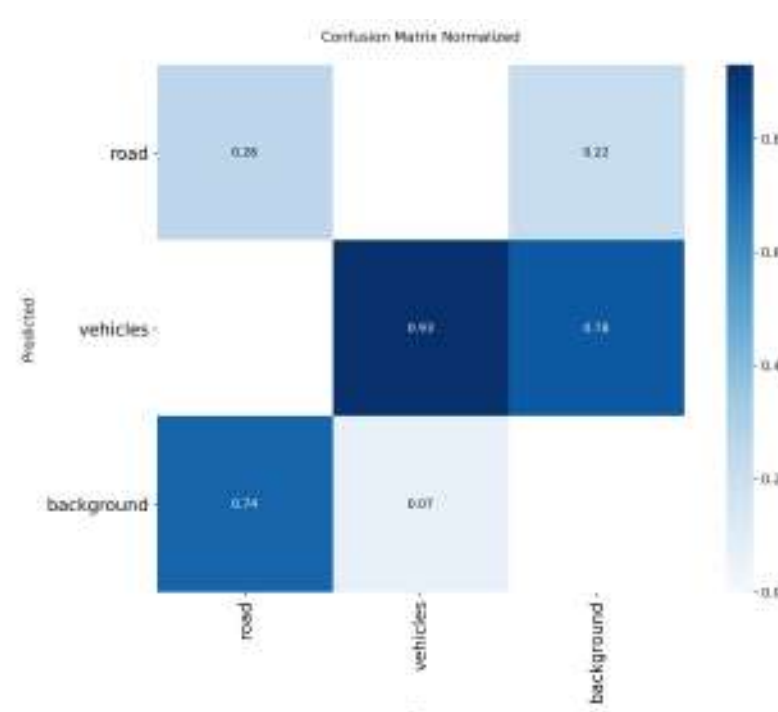

Figure 51: Confusion matrix.

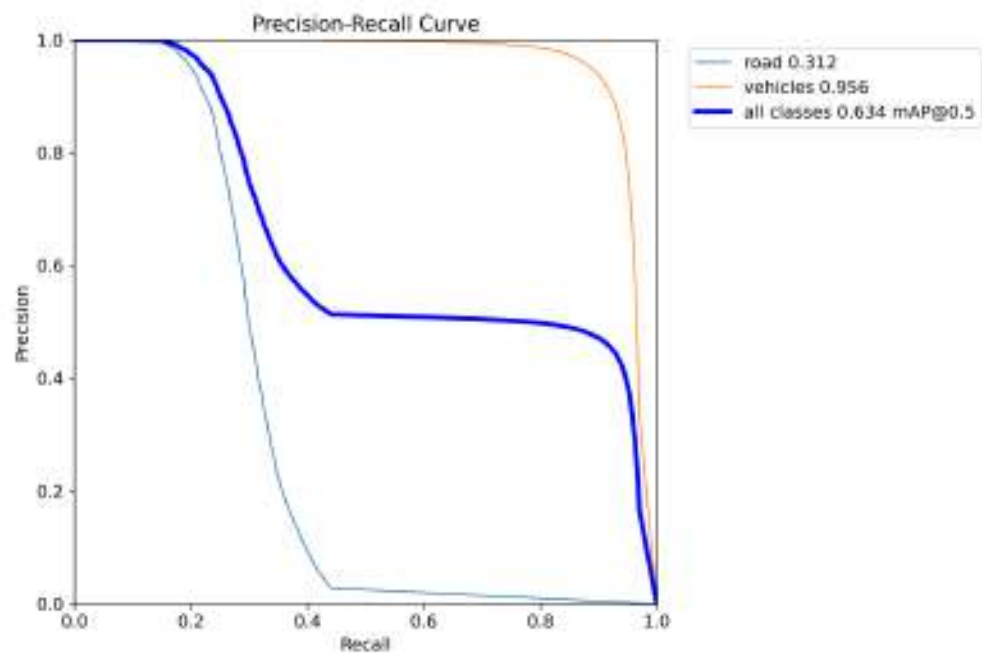

Figure 52: PR curve.

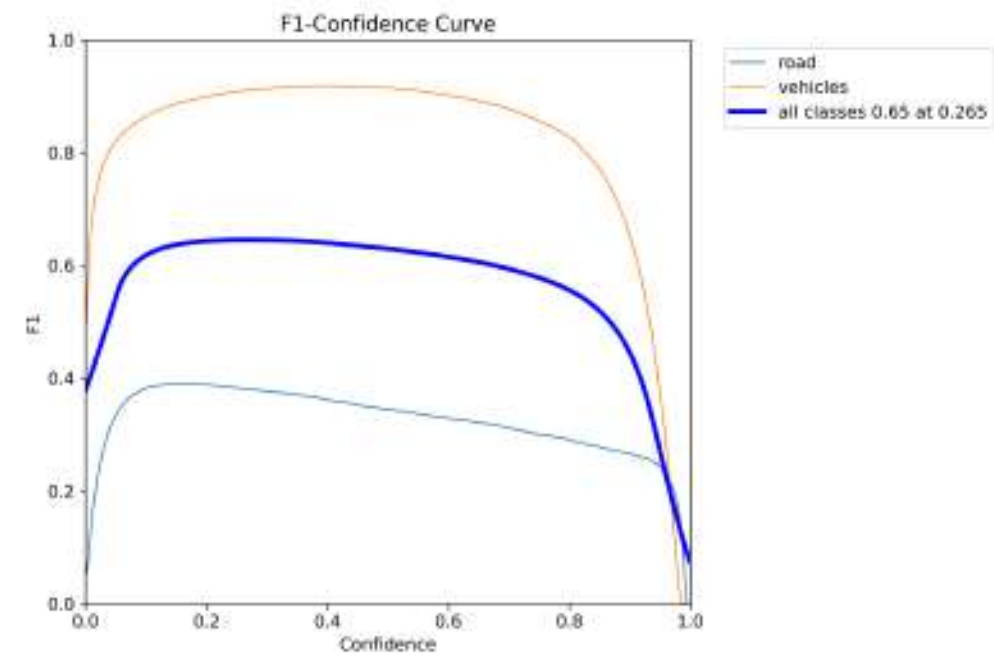

Figure 53: F1 curve.

## 8.9 YOLO11m

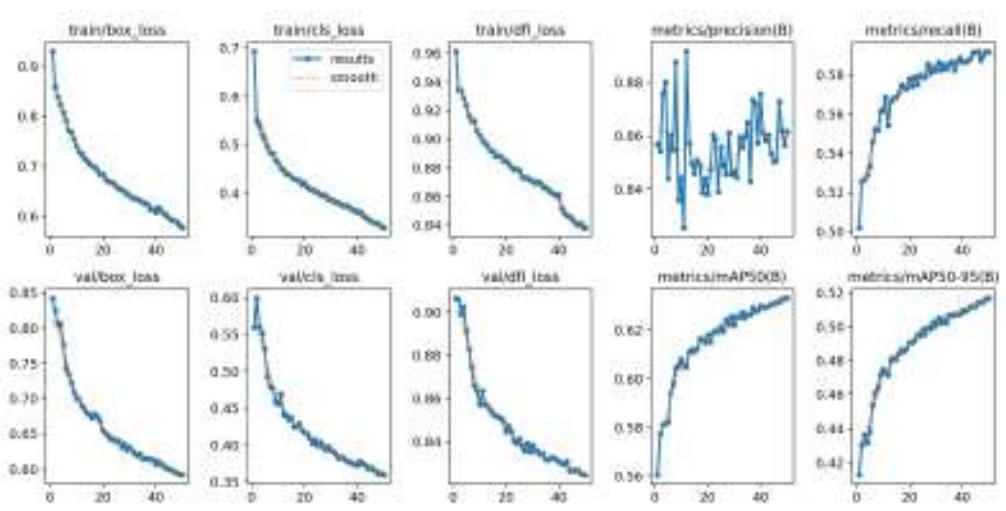

Figure 54: Training curves.

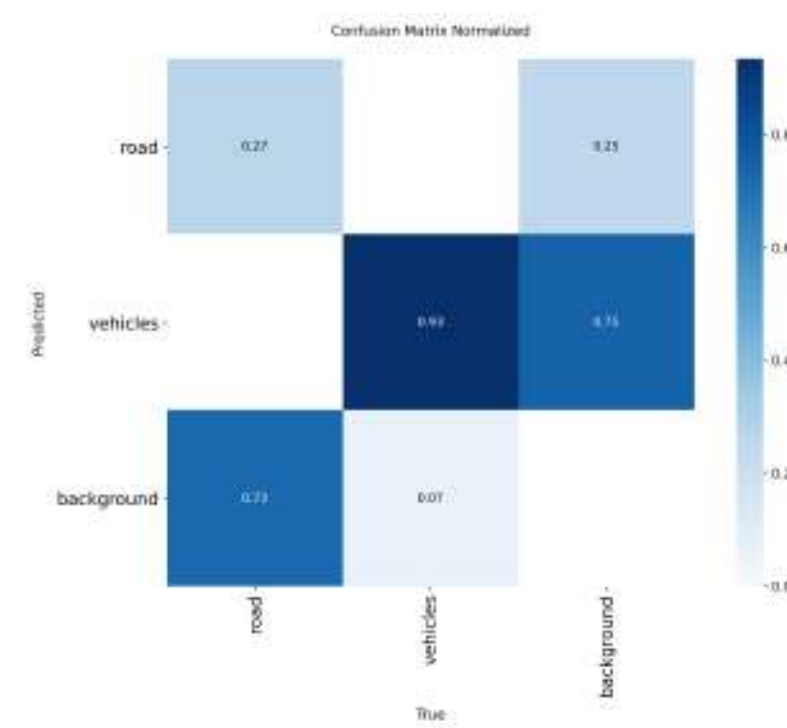

Figure 55: Confusion matrix.

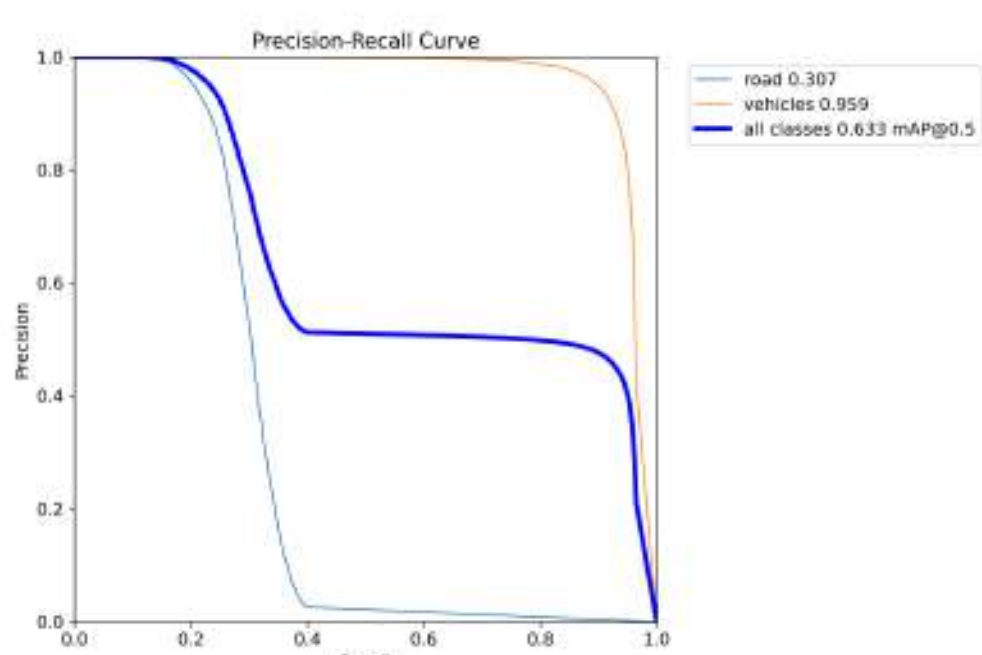

Figure 56: PR curve.

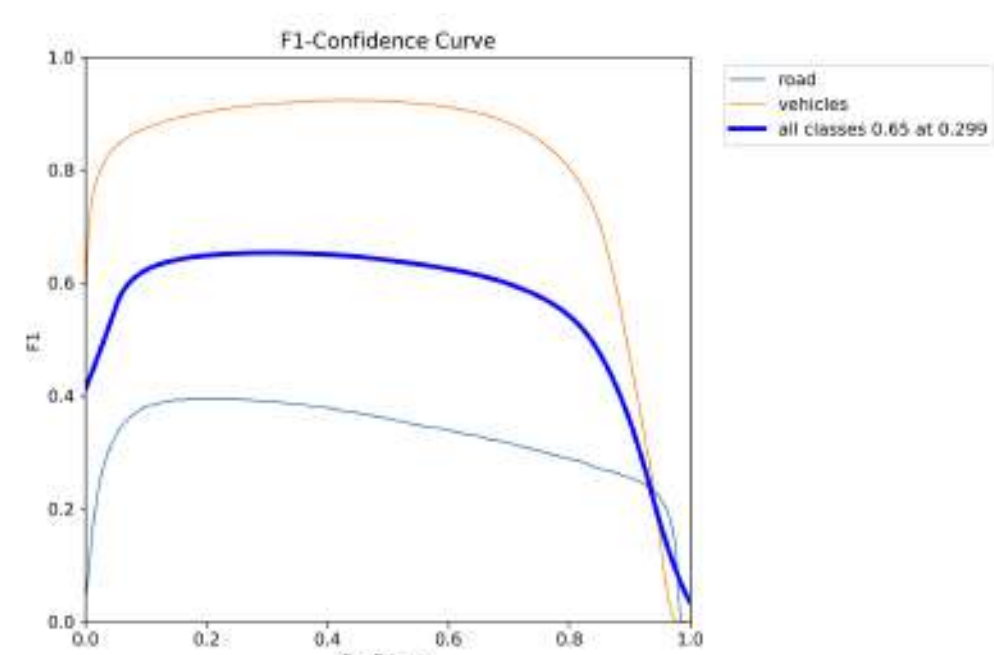

Figure 57: F1 curve.

## 8.10 RT-DETR-l

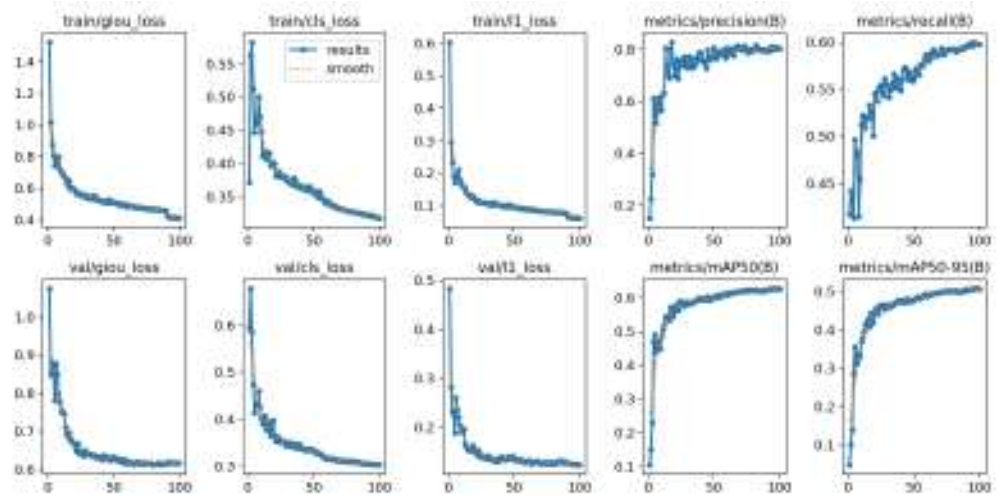

Figure 58: Training curves.

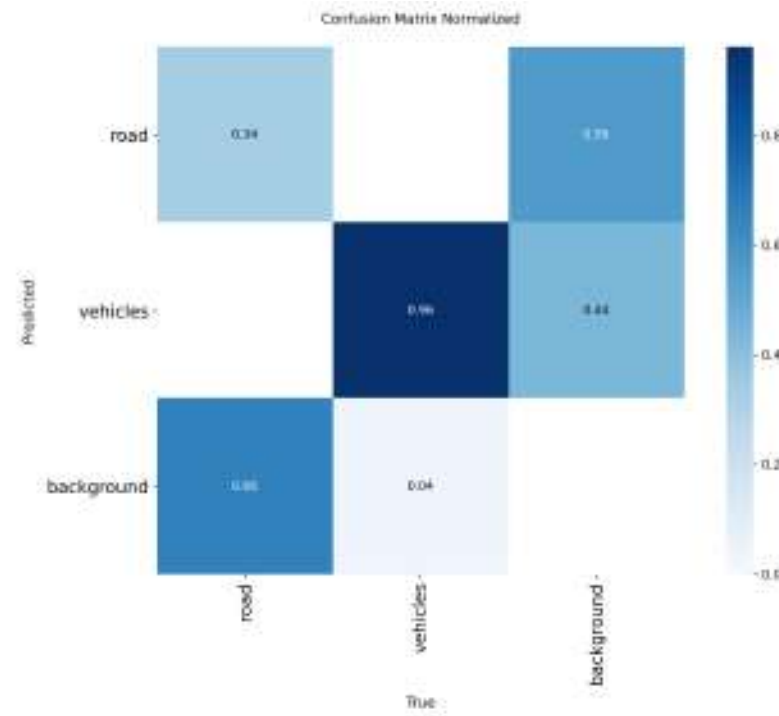

Figure 59: Confusion matrix.

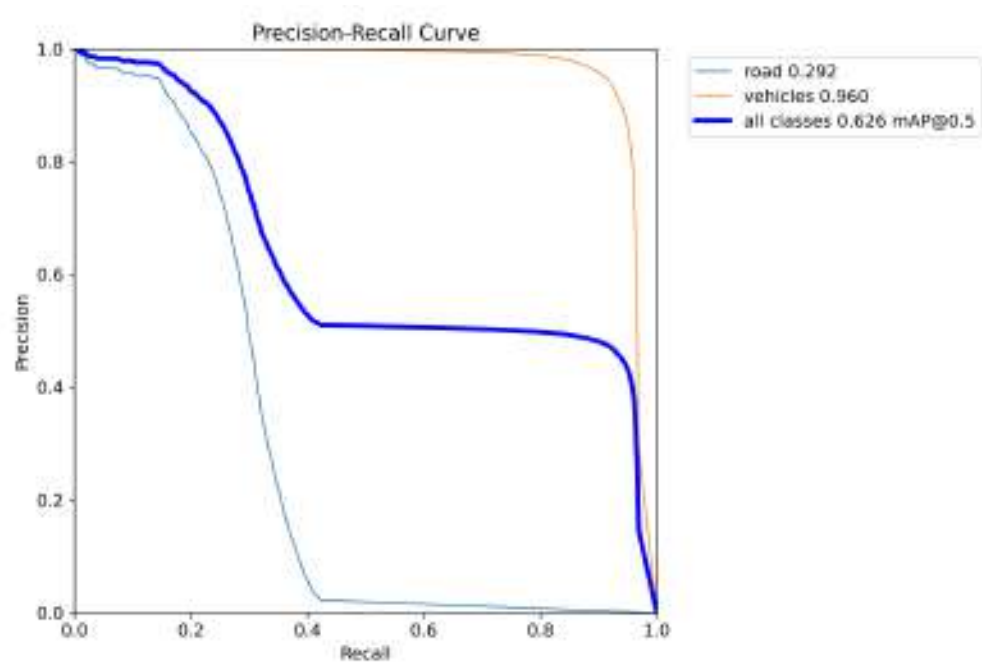

Figure 60: PR curve.

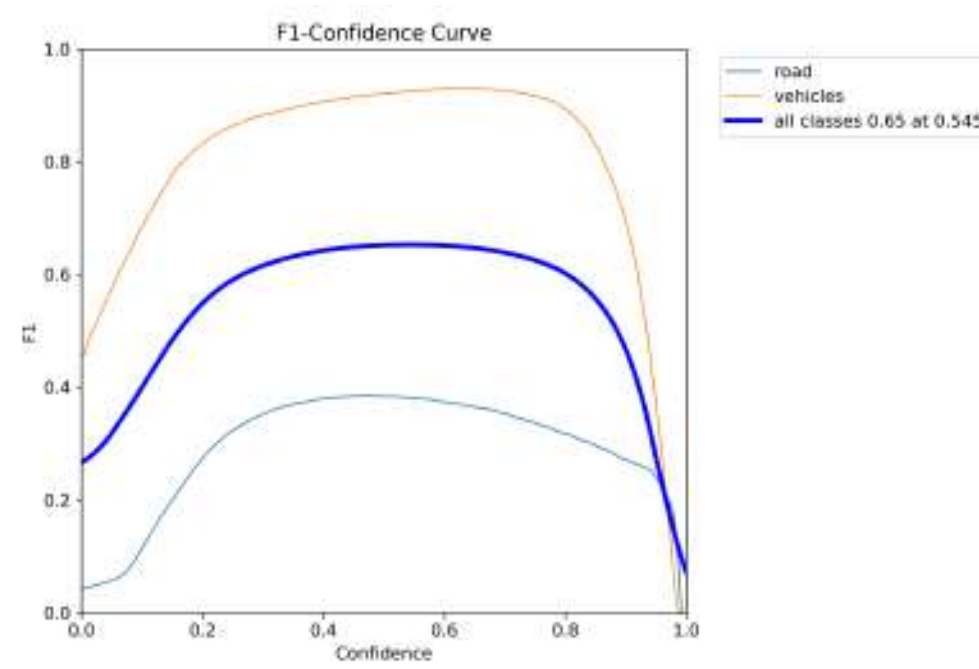

Figure 61: F1 curve.